\pdfoutput=1

\documentclass[11pt]{article}
\usepackage{hyperref}
\usepackage{emnlp2023}

\usepackage{times}
\usepackage{latexsym}
\usepackage{multirow}
\usepackage{multicol}
\usepackage[T1]{fontenc}
\usepackage[utf8]{inputenc}

\usepackage{microtype}

\usepackage{inconsolata}

\usepackage{booktabs} 
\usepackage{amsfonts}
\usepackage[T1]{fontenc}
\usepackage{adjustbox}
\usepackage{smiles}
\usepackage{bbm, dsfont}
\usepackage[normalem]{ulem}

\usepackage[utf8]{inputenc}
\usepackage[ruled,vlined]{algorithm2e}
\usepackage{enumitem}
\usepackage{subfigure}

\usepackage{colortbl}
\usepackage{pifont}%
\definecolor{maroon}{cmyk}{0,0.1,0.01,0.01}
\definecolor{blue}{cmyk}{0.95,0.0,0.2,0.2}
\definecolor{yellow}{cmyk}{0.01,0.0,0.2,0.01}
\definecolor{lightblue}{cmyk}{0.1,0.0,0.02,0.02}

\newcommand{\std}[1]{\tiny{$\pm$#1}}
\newcommand{\ours}{\textsc{ReGen}}

\newcommand{\blue}[1]{\textcolor{blue}{\small #1}}
\newcommand{\red}[1]{\textcolor{red}{\small #1}}
\newcommand{\green}[1]{\textcolor{teal}{\small #1}}
\newcommand{\orange}[1]{\textcolor{orange}{\small #1}}
\newcommand{\purple}[1]{\textcolor{magenta}{\small #1}}

\title{{\ours}: Zero-Shot Text Classification via Training Data Generation with Progressive Dense Retrieval}

\author{Yue Yu$^1$, Yuchen Zhuang$^1$, Rongzhi Zhang$^1$, Yu Meng$^2$, Jiaming Shen$^3$, Chao Zhang$^1$ \\
$^1$ Georgia Institute of Technology, GA, USA \\ $^2$ University of Illinois at Urbana-Champaign, IL, USA \\ $^3$ Google Research, NY, USA \\
\{yueyu, yczhuang, rongzhi.zhang, chaozhang\}@gatech.edu 
\\ yumeng5@illinois.edu, jmshen@google.com
}

\begin{document}
\maketitle
\begin{abstract}
With the development of large language models (LLMs), zero-shot learning has attracted much attention for various NLP tasks.
Different from prior works that generate training data with billion-scale natural language generation (NLG) models, we propose a retrieval-enhanced framework to create training data from a general-domain unlabeled corpus. 
To realize this, we first conduct contrastive pretraining to learn an unsupervised dense retriever for extracting most relevant documents using class-descriptive verbalizers. 
We then further propose two simple strategies, namely \emph{Verbalizer Augmentation with Demonstrations} and \emph{Self-consistency Guided Filtering} to improve the topic coverage of the dataset while removing noisy examples.  
Experiments on nine datasets demonstrate that {\ours} achieves 4.3\% gain over strongest baselines and saves around 70\% of the time when compared with baselines using large NLG models. 
Besides, {\ours} can be naturally integrated with recently proposed large language models to boost performance\footnote{The code and unlabeled corpus will be released in \texttt{\url{https://github.com/yueyu1030/ReGen}}.}.
\end{abstract}
\vspace{-1ex}
\section{Introduction}
Text classification serves as a fundamental task in Natural Language Processing (NLP) with a broad spectrum of applications. 
Recently, large pretrained language models (PLMs)~\cite{bert} have  achieved strong performance on text classification with a large amount of task-specific training data. 
However, in real world scenarios, collecting labeled data can be challenging due to the cost of time, money, and domain expertise. 

To reduce the burden of human annotation, we study automatic \emph{dataset generation} for text classification under the zero-shot setting, where no \emph{task-specific} or \emph{cross-task} data is available. 
Such a setting is different from previous works that use a large collection of labels from auxiliary tasks for zero-shot text classification~\cite{yin-etal-2019-benchmarking,gera2022zero,flan,t0},
and is particularly challenging since we need to adapt the language understanding abilities of PLMs to target classification tasks with minimal supervision. 

Prior works on zero-shot synthetic dataset generation mainly fall into two categories: 
(1) \emph{Generative methods} leverage a billion-scale NLG model to generate class-conditioned texts for PLM fine-tuning~\cite{meng2022generating,ye2022progen,ye2022zerogen}. While these methods work well on easy tasks (\eg binary classification), they can be fragile on challenging tasks with more classes, as the generated text can be less discriminative. Besides, the gigantic size of the NLG model 
will also cause the inefficiency issue.  
(2) \emph{Mining-based} methods
design rule-based regular expressions to extract text from the background corpus as synthesized training data~\cite{van2022don}, but these rules are often too simple to capture the complex semantics of text. As a result, the mined dataset contains many incorrectly-labeled data, and the fine-tuned PLM can easily overfit noisy labels. 

We design a new framework {\ours}\footnote{\underline{\textbf{R}}etrieval-\underline{\textbf{E}}nhanced Zero-shot Data \underline{\textbf{Gen}}eration.} to solve zero-shot text classification. 
The setting of {\ours} is close to the mining-based technique~\cite{van2022don}, where a set of class-specific verbalizers and a collection of general-domain unlabeled corpus are available. 
Motivated by the limitation of hard matching with regular expressions which hardly preserves the meaning of verbalizers, we propose to leverage \emph{dense retrieval} (DR)~\cite{lee-etal-2019-latent,karpukhin-etal-2020-dense,ance,sun2022reduce,cui2022can}, which calculates semantic relevance in a continuous representation space, for dataset curation. 
With such a \emph{soft matching} mechanism, DR is able to better encode the category-specific semantics and thus fetch the relevant documents from the corpus. 
To integrate DR with the target classification task, we employ two PLMs: one retrieval model $(R_{\theta})$ to extract the most relevant documents from the unlabeled corpus for synthetic dataset curation, and one classification model $(C_{\phi})$ to be fine-tuned on the generated synthetic dataset to perform the downstream task.  
Before performing text retrieval, we first conduct contrastive learning on the unlabeled corpus to further pretrain the retrieval model $R_{\theta}$ for producing better sequence embeddings. 
Then, with the retrieval model, we use the verbalizers from each class as queries to  retrieve relevant documents from the unlabeled corpus, which will be used as the training data for target tasks.

Simply fine-tuning the classifier on the above training data may yield limited performance, as the verbalizers are often too generic to cover all the category-related topics (\eg, the word `sports' alone does not cover concrete types of sports). Thus, the retrieved data may contain noisy and irrelevant documents. 
To enhance the quality of the synthetic dataset, we conduct multi-step retrieval with two additional strategies to strengthen our framework:
(1) we \emph{augment} the verbalizer with the retrieved documents from the previous round as additional information~\cite{prf} to enrich its representation, which allows for extracting more relevant documents for the downstream task. 
(2) we exploit \emph{self-consistency} to filter the potentially incorrect examples  
when the pseudo labels produced by the retrieval model ($R_{\theta}$) and the classifier ($C_{\phi}$) disagree with each other. 
We note that {\ours} \emph{does not} use annotated labels from any other tasks, making it applicable to the true zero-shot learning. 
Besides, {\ours} only requires two BERT$_\text{base}$ scale PLMs, which is  efficient compared with methods using large NLG models.

Our contribution can be summarized as follows: (1) We propose {\ours}, a framework for zero-shot dataset generation with a general-domain corpus and  retrieval-enhanced language models. 
(2) We develop two additional techniques, namely verbalizer augmentation with demonstration and self-consistency guided filtering to improve the quality of the synthetic dataset.  
(3) We evaluate {\ours} on \emph{nine} NLP classification tasks to verify its efficacy. We also conduct detailed analysis to justify the role of different components as well as the robustness of {\ours} over different verbalizers.





\vspace{-0.4ex}
\section{Related Work}
\vspace{-0.4ex}
Zero-shot Text Classification (ZSTC) aims to categorize the text document without using task-specific labeled data. 
With pretrained language models, a plenty of works attempted to convert the classification task into other formats such as masked language modeling~\cite{hu-etal-2022-knowledgeable,gao-etal-2021-making}, question answering~\cite{zhong-etal-2021-adapting-language,flan,t0} or entailment~\cite{yin-etal-2019-benchmarking,gera2022zero}  
for zero-shot learning. These works are orthogonal to {\ours} as we do not directly perform inference and do not leverage human annotations from additional tasks.

More relevant to us, there are some recent studies that perform ZSTC via generating a task-specific dataset using NLG models, which is then used to fine-tune a classifier for the target task such as text  classification~\citep{ye2022progen,ye2022zerogen,meng2022generating}, sentence similarity calculation~\citep{schick-schutze-2021-generating}, commonsense reasoning~\cite{yang-etal-2020-generative,kan2021zero}, and instruction-based tuning~\cite{wang2022self}. Unfortunately, the generation step is time-consuming and the quality of the generated text can be less satisfactory in capturing fine-grained semantics.  
The most relevant work to us is \citep{van2022don}, which also extracts documents from the unlabeled corpus to form the training set. But it simply uses regular expressions to mine documents and cannot fully capture the contextual information of verbalizers. 
Instead, we leverage dense retrieval for concept understanding and obtain the most relevant documents, which is combined with verbalizer augmentation to improve retrieval quality.

On the other hand, retrieval-augmented language models have been used in language modeling~\cite{Khandelwal2020Generalization,pmlr-v162-borgeaud22a}, OpenQA~\cite{jiang2022retrieval,sachan2021endtoend},  information extraction~\cite{zhuang-etal-2022-resel} and knowledge-intensive tasks~\cite{lewis2020retrieval,izacard2022few},  where tokens or documents are retrieved based on contextual representations and are used as additional inputs to support target tasks. While such a paradigm has also been explored for zero-shot learning, it is mainly used for  zero-shot prompt-based inference~\cite{shi2022nearest,chen2022decoupling}. 
Instead, we empirically demonstrate the efficacy of retrieval-enhanced learning for zero-shot dataset curation with an unsupervised dense retrieval model.




\vspace{-0.5ex}
\section{Preliminaries}
\vspace{-0.5ex}
\noindent  $\diamond$ \textbf{Setup.} 
We focus on synthesizing a task-specific dataset for text classification~\cite{meng2022generating,van2022don}. 
Besides, we stick to the \emph{strict} zero-shot  setup~\cite{perez2021true}, where \emph{no labeled examples} from either target tasks or other tasks are available. 


\noindent  $\diamond$ \textbf{Available Resources.}
Besides annotated labels, the availability of {massive task-specific unlabeled data} is also a rarity --- in prior works, such unlabeled data is obtained via removing the ground-truth label from the original dataset~\cite{meng-etal-2020-text}, and can be scarce in real zero-shot settings~\cite{tam-etal-2021-improving}. 
The most accessible information is a collection of general-domain unlabeled corpus $\mathcal{D}$ (\eg, \textsc{Wiki}), which is freely available online and has been used for  pretraining~\cite{bert,gururangan-etal-2020-dont}. 
Recent works have also use such an external corpus for zero-shot learning~\cite{shi2022nearest,van2022don}.

\noindent  $\diamond$ \textbf{Task Formulation.} With the above discussion, we consider the classification task where we are given the label set $\mathcal{Y}=\{1, 2, \ldots, c\}$ ($c$ is the number of classes), and a mapping $\mathcal{M}:\mathcal{Y}\to\mathcal{W}$ that converts each label $y \in \mathcal{Y}$ into a class-descriptive verbalizer $w_y \in \mathcal{W}$. We also assume a general-domain unlabeled corpus $\cD$ is available. 
We seek to curate training data $\cT$ from $\cD$  
and learn a PLM $C_{\phi}$ which will be fine-tuned as the classifier.

\noindent  $\diamond$ \textbf{Backgrounds for Dense Retrieval (DR).} In dense retrieval~\cite{lee-etal-2019-latent}, the PLM is used to represent queries and documents in dense vectors. The relevance score $f(q, d)$ is calculated with a scoring function (\eg, dot product) between query and document vectors
\begin{equation}
\setlength{\abovedisplayskip}{5.2pt}
\setlength{\belowdisplayskip}{5.2pt}
f(q, d)=\operatorname{sim}\left({R}_{\theta}(q), {R}_{\theta}(d)\right),
\label{eq:sim}
\end{equation}
where the embedding of the $\texttt{[CLS]}$ token from the final layer of $R_{\theta}$ is used as the representation for both queries and documents. 
In practice, the documents are encoded offline, and can be efficiently retrieved using approximate nearest neighbor search (ANN)  with the queries~\cite{faiss}.

\vspace{-0.3ex}
\section{Method}
\vspace{-0.3ex}
In this section, we present {\ours} (our framework) and introduce the major components. 
\subsection{Contrastive Pretraining for Retriever $R_{\theta}$} 
\label{sec:coco}
Directly using BERT for retrieval can lead to unsatisfactory results since BERT embeddings are not tailored for retrieval application~\cite{gao-etal-2021-simcse}. 
To effectively train a dense retrieval model \emph{without relevance supervision}, we hypothesize that two sentences from the same document share similar semantics as they may describe the same topic. Then, we  continuously pretrain the PLM on the corpus $\cD$ with contrastive learning~\cite{gao-callan-2022-unsupervised,izacard2021contriever,yu2022cocodr}: 
Given a document $d_i\in\cD$, 
 the positive pair $(x_i, x_i^{+})$ is constructed by randomly sampling two disjoint sentences from $d_i$.
Let $\mathbf{h}_i=R_{\theta}(x_i), \mathbf{h}_i^{+}=R_{\theta}(x_i^{+})$ denote the representation of $x_i$ and $x_i^{+}$ encoded by the retriever $R_{\theta}$, the training objective of contrastive learning for pair $(x_i, x_i^{+})$ with  a mini-batch of $N$ pairs is: 
\begin{equation}
\setlength{\abovedisplayskip}{5.2pt}
\setlength{\belowdisplayskip}{5.2pt}
\ell_{\text{cl}}=-\log \frac{{e}^{\operatorname{sim}\left(\mathbf{h}_i, \mathbf{h}_i^{+}\right) / \tau}}{\sum_{j=1}^N {e}^{\operatorname{sim}\left(\mathbf{h}_i, \mathbf{h}_j^{+}\right) / \tau}},
\label{eq:cl}
\end{equation}
where we use in-batch instances as negative samples~\cite{gillick-etal-2019-learning}, $\operatorname{sim}(\mathbf{h}_i, \mathbf{h}_i^{+})=\mathbf{h}_i^{\top}\mathbf{h}_i^{+}$ is the dot product, and $\tau=1$ is the parameter for temperature. 
Contrastive learning improves the representations by promoting the alignment of similar text sequences and the uniformity of unrelated text sequences, thus enhancing the embedding quality for documents in $\cD$.
\vspace{-0.3ex}
\subsection{Overall Pipeline}
\vspace{-0.3ex}
\label{sec:pipeline}
With a pretrained retrieval model $R_{\theta}$, {\ours} follows a \emph{retrieve-then-finetune} pipeline to curate the training data from the  corpus $\cD$ which will be used to finetune the PLM classifier $C_{\phi}$.  
The details of our framework are described as follows.

\noindent \textbf{Document Retrieval with Verbalizers.} 
With the class-specific verbalizers, we construct the input queries for each class to retrieve the relevant documents from $\cD$. Formally, the query for the $i$-th class $(1\leq i \leq c)$ can be expressed as 
\begin{equation}
\setlength{\abovedisplayskip}{5pt}
\setlength{\belowdisplayskip}{5pt}
\label{eq:query}
q_i = \texttt{[CLS]} \circ \cP(w_i) \circ \texttt{[SEP]}, \nonumber
\end{equation}
where $\cP(w_i)$ is the template for the corresponding class with the verbalizer $w_i$ and $\circ$ stands for the concatenation operation. For instance, a query for the binary sentiment classification can be formulated as \underline{$q_i = \texttt{[CLS]} \ \ \text{It was} \ w_{i} \ \ \texttt{[SEP]}$}, 
where $w_1$ and $w_2$ $(c=2 \ \text{in this case})$ stand for the  verbalizers, namely ``\emph{bad}'' (negative) and  ``\emph{great}'' (positive), respectively. 
By feeding the class-dependent query into the retriever $R_{\theta}$, we expect the retriever to understand its contextualized semantics~\citep{rubin-etal-2022-learning}, and extract the relevant documents from the corpus which serve as training examples for the corresponding category.  
For the $i$-th class, the initial retrieved dataset $\cT_i^{1} \subset \cD$ can be written as 
\begin{equation}
\setlength{\abovedisplayskip}{5pt}
\setlength{\belowdisplayskip}{5pt}
\cT_i^{1}=\underset{d\in \cD}{\operatorname{Top-k}} \ f(q_i, d),
\label{eq:knn}
\end{equation}
where $f(q, d)$ is defined in Eq.~\ref{eq:sim}. The full retrieved dataset can be expressed as $\cT^{1}=\underset{1\leq i \leq c}{\cup} \cT_i^{1}$.

\noindent \textbf{Fine-Tuning PLM with Curated Data.} 
After obtaining the training data $\cT$ from the corpus\footnote{Here we omit the superscript for $\cT$ as the fine-tuning procedure remains the same for all rounds and generated datasets.}, one can fine-tune a PLM classifier $C_{\phi}$ for the downstream task. 
To achieve better fine-tuning stability and generalization, we adopt the simple \emph{label smoothing} (LS) technique~\cite{muller2019does}, which mixes the one-hot 
labels with uniform vectors. 
For a training example $(x, y) \in \cT$, $C_{\phi}$ is trained to minimize the divergence between the label and the classifier's prediction $p_{\phi}(x)$ as 
\begin{equation}
\setlength{\abovedisplayskip}{5pt}
\setlength{\belowdisplayskip}{5pt}
\label{eq:label_smooth}
\min_{\phi} \ \ \ell_{\text{ft}}=-\sum_{j=1}^{c} q_j \log(p_{\phi}(x)_j), 
\end{equation}
where $q_j = \mathbbm{1}(j=y)(1-\alpha) + {\alpha}/{c}$ is the smoothed label and $\alpha=0.1$ is the smoothing term.
LS prevents $C_{\phi}$ from overfitting to training data by forcing it to produce less confident predictions.

\begin{algorithm}[!t]
	\begin{small}
	\KwIn{$\cD$: Unlabeled Corpus; $\mathcal{Y}$: Label space; $\mathcal{P}$: Verbalizers; $R_{\theta}$: Retrieval Model; $C_{\phi}$: Classification Model; $T$: Rounds of Retrieval.}
	\blue{// Step 0: \textit{Contrastive Learning.}} \\ 
    Pretrain $R_{\theta}$ with Contrastive Learning via Eq.~\ref{eq:cl}.  \\
	\For{$t = 1, 2, \cdots, T$}{ 
		{
		   \blue{// {Step 1}: \textit{(Multi-step) Document Retrieval.}} \\
		    \If{$t=1$}{
              Retrieve Documents $\cT^1$ with $\cP$ via Eq.~\ref{eq:knn}. 
           }\Else{
              Retrieve Documents $\cT^t$ with $\cP$ and $\tilde{\cT}^{t-1}$  via Eq.\ref{eq:knn_expansion}. \blue{// \textit{Verbalizer Augmentation.}} 
                 }  
            \blue{// Step 2: \textit{Document Filtering.}} \\
            Obtain Filtered Dataset $\tilde{\cT}^t$ via Eq.~\ref{eq:filter}. \\
            \blue{// Step 3: \textit{Language Model Fine-tuning.}} \\
            Fine-tune PLM $C_{\phi}^{t}$ with $\tilde{\cT}^t$ via Eq.~\ref{eq:label_smooth}.
		}
	}
	\KwOut{The dataset $\tilde{\cT}^t$ and the PLM classifier $C_{\phi}^{t}$.}
	\end{small}
	\caption{Process of {\ours}. }
	\label{alg:main}
\end{algorithm}
\vspace{-0.3ex}
\subsection{Progressive Training Data Curation via Multi-step Dense Retrieval} 
Although the aforementioned pipeline can retrieve a set of documents used for training ($\cT^1$), the performance can still be suboptimal because (1) the training set only have \emph{limited coverage} as the verbalizers only contains few key words which is too specific to fully represent the categorical information. 
(2) the training set still contain \emph{noisy} or \emph{task-irrelevant} documents as the $R_{\theta}$ may not always retrieve texts pertaining to the desired class. 
To overcome these drawbacks, we perform  document retrieval for multiple rounds, employing two additional strategies as described below.
\paragraph{Verbalizer Augmentation with Demonstrations.} The verbalizers often contain only a few words and are insufficient to perfectly reflect the underlying information.  Motivated by the recently proposed demonstration-based learning~\cite{gpt3,min2022rethinking} which augments the input with labeled examples to support in-context learning, we aim to enrich verbalizers with top retrieved documents for improving their representations~\cite{ance-prf}, and thus enhancing the quality of the 
retrieved data.  
Specifically, in the $t$-th ($t>1$) round, we use the retrieved documents from the $t$-1 round as demonstrations to augment the verbalizer for the $i$-th class as\footnote{We obtain \emph{multiple} queries for each class after this step.}
\begin{equation}
\setlength{\abovedisplayskip}{5pt}
\setlength{\belowdisplayskip}{5pt}
q_{i,j}^{t} = \texttt{[CLS]} \circ \cP(w_i) \circ \texttt{[SEP]} \circ d_{i, j}^{t-1} \circ \texttt{[SEP]},  
\label{eq:query_expansion}
\end{equation}
where $d_{i,j}^{t-1}$ is the $j$-th documents for the $i$-th class in the previous dataset $\tilde{\cT}^{t-1}$.  
With the augmented queries, 
$\cT_i^{t}$ and $\cT^t$ are obtained via combining the retrieved documents  as 
\begin{equation}
\setlength{\abovedisplayskip}{5pt}
\setlength{\belowdisplayskip}{5pt}
\cT_i^{t}=\underset{j}{\bigcup} \ (\underset{d\in \cD}{\operatorname{Top-k}} \ f(q_{i,j}^t, d)), \cT^{t}=\underset{1\leq i \leq c}{\cup} \cT_i^{t}.
\label{eq:knn_expansion}
\end{equation}
\paragraph{Filtering Noisy Data guided via Self-consistency.}
The above retrieval process may also introduce noisy examples due to the limited capability of the retrieval model. 
While the label smoothing in Eq.~\ref{eq:label_smooth} can mitigate this issue during fine-tuning, it is a generic technique without considering task-specific knowledge. 
To further fulfill the denoising purpose, we simply leverage the classifier from the previous round and exploit the \emph{consistency} between the retriever and classifier to identify potential incorrect examples. 
For the example from the $t$-th round $(t>1)$ denoted as $(x^t, y^t) \in \cT^{t}$ where $y^t$ is the label for the augmented verbalizer, we generate the predicted label using the classifier $C_{\phi}^{t-1}$ from the previous round\footnote{When $t=1$, we use the zero-shot prompting model as the classifier due to the absence of the `previous model'.}
 as 
$\hat{y}^{t-1}=\argmax p_{\phi}^{t-1}(x^t)$. 

\noindent Then, the filtered dataset $\tilde{\cT}^{t}$ is expressed as 
\begin{small}
\begin{equation}
\setlength{\abovedisplayskip}{5pt}
\setlength{\belowdisplayskip}{5pt}
\tilde{\cT}^{t} = \{(x^t,y^t) \in {\cT}^{t} \mid \argmax p_{\phi}^{t-1}(x^t)=y^t \}.
\label{eq:filter}
\end{equation}
\end{small}
To interpret Eq.~\ref{eq:filter}, we only preserve examples where the prediction from the previous classifier $\hat{y}^{t-1}$ and the retrieved label $y^t$ are \emph{consistent} to fine-tune the classifier $C_{\phi}$, thus serving as an additional protection for $C_{\phi}$ against overfitting to label noises.

\begin{table}[t]
\centering
	\resizebox{\columnwidth}{!}{
		\begin{tabular}{ccccc}
			\toprule \bf Dataset & \bf Task  & \bf Class  & \bf \# Test & \bf Metric  \\ \midrule
            AGNews & News Topic &4  & 7.6k & Accuracy \\ 
            DBPedia & Wikipedia Topic & 14 & 70k & Accuracy  \\
            Yahoo Topics & Web QA Topic & 10 & 60k & Accuracy  \\
            NYT & News Topic & 9 & 30k & F1  \\\hline
            IMDB & Movie Review Sentiment &2  & 25k & Accuracy  \\ 
            MR & Movie Review Sentiment& 2& 2k & Accuracy  \\ 
            SST-2 & Movie Review Sentiment& 2 & 0.8k & Accuracy \\ 
            Amazon & Product Review Sentiment& 2 & 40k & Accuracy \\ 
            Yelp &Restaurant Review Sentiment& 2 & 38k & Accuracy \\ 
            \bottomrule
		\end{tabular}
	}
	\caption{Dataset statistics. }
	\label{tab:dataset}
\end{table}

\begin{table*}[!t]
\vspace{-0.3ex}
\centering 
\renewcommand\arraystretch{0.9}
\fontsize{7.5}{9.5}\selectfont \setlength{\tabcolsep}{0.5em}
\resizebox{0.99\linewidth}{!}{%
\begin{tabular}{l|ccccc|cccccc|c}
\toprule
\bf    Task ($\rightarrow$)     & \multicolumn{5}{c|}{\textbf{Topic Classification}}  &  \multicolumn{6}{c|}{\textbf{Sentiment Classification}} & \bf All  \\  \hline   
\bf    Method ($\downarrow$) / Dataset ($\rightarrow$)    & \bf  AG News & \bf DBPedia & \bf Yahoo  & \bf NYT & \multicolumn{1}{>{\columncolor{maroon}}c}{\bf Avg.} & \bf  IMDB & \bf MR & \bf SST-2 & \bf Amazon  & \bf Yelp &  \multicolumn{1}{>{\columncolor{lightblue}}c}{\bf Avg.} & \multicolumn{1}{>{\columncolor{yellow}}c}{\bf Avg.} \\ \hline
\multicolumn{10}{l}{\emph{Zero-shot Learning via Direct Inferencing on Test Data}}  \\\hline 
NSP-BERT~\shortcite{sun2021nsp} & 78.1 & 69.4 & 47.0 & 54.6 & \multicolumn{1}{>{\columncolor{maroon}}c}{ 62.3} & 73.1 & 74.4 & 75.6 & 69.4 & 66.3 &  \multicolumn{1}{>{\columncolor{lightblue}}c}{71.8}& \multicolumn{1}{>{\columncolor{yellow}}c}{67.5} \\ 
Prompt~\shortcite{schick-schutze-2021-exploiting} & 73.2 & 71.3 & 44.1 & 57.4 & \multicolumn{1}{>{\columncolor{maroon}}c}{61.5}  &  74.8 & 73.2 & 75.9 & 80.2 & 78.1 & \multicolumn{1}{>{\columncolor{lightblue}}c}{76.4}  & \multicolumn{1}{>{\columncolor{yellow}}c}{68.9}  \\
KNN-Prompt~\shortcite{shi2022nearest} & 78.8 & --- & 51.0 & --- & \multicolumn{1}{>{\columncolor{maroon}}c}{---}  &  --- & 78.2 & 84.2 & 85.7 & --- & \multicolumn{1}{>{\columncolor{lightblue}}c}{---} & \multicolumn{1}{>{\columncolor{yellow}}c}{---} \\
GPT-3$^\ddagger$~\shortcite{zhao2021calibrate} & 73.9 & 59.7 & 54.7 & 57.0 & \multicolumn{1}{>{\columncolor{maroon}}c}{61.3}  & 75.8 & 76.3 & 87.2 & 75.0 & 78.5 & \multicolumn{1}{>{\columncolor{lightblue}}c}{78.6} & \multicolumn{1}{>{\columncolor{yellow}}c}{69.9} \\
\hline
\multicolumn{10}{l}{\emph{Zero-shot Learning via Generating Task-specific Datasets}} 
 \\\hline
SuperGen$^\ddagger$~\shortcite{meng2022generating} & 77.4\std{1.5} & 66.5\std{2.0} & 40.8\std{1.5}& 53.9\std{1.5} & \multicolumn{1}{>{\columncolor{maroon}}c}{59.7} & 85.8\std{1.6} & 81.9\std{0.9} & 88.6\std{0.5} & 91.0\std{0.9} & \bf  93.6\std{0.6} &  \multicolumn{1}{>{\columncolor{lightblue}}c}{88.1} & \multicolumn{1}{>{\columncolor{yellow}}c}{73.9}
\\
Mining$^*$~\shortcite{van2022don} & 79.2 & 80.4 & 56.1 & --- & \multicolumn{1}{>{\columncolor{maroon}}c}{---} & 86.7 & 80.5 & 85.6 & 92.0 & 92.0 & \multicolumn{1}{>{\columncolor{lightblue}}c}{87.3} & \multicolumn{1}{>{\columncolor{yellow}}c}{---}\\
Mining$^{*\natural}$ (\emph{Our ReImp.}) & 79.7\std{1.0} & 82.1\std{0.6} & 57.0\std{0.6} & 68.6\std{0.9} & \multicolumn{1}{>{\columncolor{maroon}}c}{71.9}  &
87.1\std{0.6}  & 79.9\std{0.7} &  85.0\std{0.6} & 92.1\std{0.5} & 92.3\std{0.5} & \multicolumn{1}{>{\columncolor{lightblue}}c}{87.2} & \multicolumn{1}{>{\columncolor{yellow}}c}{79.6}\\
\bf {\ours}  (Our Method) & \bf{85.0\std{0.8}} & \bf 87.6\std{0.9}& \bf  59.4\std{0.8} & \bf  74.5\std{1.1} & \multicolumn{1}{>{\columncolor{maroon}}c}{\bf 76.6}  & \bf  89.9\std{0.5} & \bf  82.5\std{0.7} & \bf  88.9\std{0.4} & \bf  92.3\std{0.4} &  93.0\std{0.5} & \multicolumn{1}{>{\columncolor{lightblue}}c}{\bf 89.3} & \multicolumn{1}{>{\columncolor{yellow}}c}{\bf 83.0}\\
\hline
\multicolumn{10}{l}{\emph{\textbf{For Reference Only}: Using labeled data from other tasks / task-specific corpus / external knowledge base.}}  
\\ \hline
TE-NLI (Best)$^\dagger$~\shortcite{yin-etal-2019-benchmarking} & 78.0 & 73.0 & 43.8 & 70.7 & \multicolumn{1}{>{\columncolor{maroon}}c}{66.4}  & 64.6 & 68.3  & 68.6 & 76.7 & 73.5 &  \multicolumn{1}{>{\columncolor{lightblue}}c}{70.3} & \multicolumn{1}{>{\columncolor{yellow}}c}{68.6} \\
NLI-ST $^{\dagger\sharp}$~\shortcite{gera2022zero} & 76.5 & 92.2 & 59.8 & --- & \multicolumn{1}{>{\columncolor{maroon}}c}{---}  & 92.5 & --- & --- & 94.3 & --- &  \multicolumn{1}{>{\columncolor{lightblue}}c}{---} & \multicolumn{1}{>{\columncolor{yellow}}c}{---} \\
KPT$^{\sharp,\S}$~\shortcite{hu-etal-2022-knowledgeable} & 84.8 & 82.2  & 61.6 & 72.1 &\multicolumn{1}{>{\columncolor{maroon}}c}{75.2}  & 91.2 & --- & --- & 92.8 & --- &  \multicolumn{1}{>{\columncolor{lightblue}}c}{---} & \multicolumn{1}{>{\columncolor{yellow}}c}{---} \\
LOTClass$^\sharp$~\shortcite{meng-etal-2020-text} & 86.2 & 91.1 & 55.7 & 49.5 & \multicolumn{1}{>{\columncolor{maroon}}c}{70.7}  & 86.5 & 70.8 & 80.9 & 91.7 & 87.6 & \multicolumn{1}{>{\columncolor{lightblue}}c}{83.5} & \multicolumn{1}{>{\columncolor{yellow}}c}{77.1} \\
X-Class$^\sharp$~\shortcite{wang-etal-2021-x} & 85.7 & 91.3 & 50.5 & 68.5 & \multicolumn{1}{>{\columncolor{maroon}}c}{74.0}  & 89.0 & 78.8 & 84.8 & 90.4 & 90.0 & \multicolumn{1}{>{\columncolor{lightblue}}c}{86.5} & \multicolumn{1}{>{\columncolor{yellow}}c}{80.3} \\
\bottomrule
\end{tabular}
}
\vspace{-1ex}
\caption{Main results. We report average performance and standard deviation across 5 runs \emph{if fine-tuning is applied}. $*$: concurrent work, $\natural$: use the same corpus and template as {\ours} for \emph{fair comparisons}, $\dagger$: use labeled data from auxiliary tasks, $\sharp$: use task-specific corpus,  $\ddagger$: use billion-scale PLMs, $\S$: use additional knowledge base.}
\label{tab:main_f1}
\vspace{-1ex}
\end{table*}

\subsection{Overall Algorithm}
The procedure of {\ours} is summarized in Algorithm~\ref{alg:main}. Note that the retrieval model pretraining and corpus indexing only need to be done \emph{once} before applying to all datasets.  
In each round of retrieval, it only needs one extra ANN retrieval operation per query, which is efficiently supported by FAISS~\cite{faiss}. We conduct the  efficiency study in the Section \ref{sec:efficiency}.
\vspace{-0.3ex}
\section{Experiments}
\vspace{-0.5ex}
\subsection{Experimental Setups}
\noindent  $\diamond$ \textbf{Datasets.} In this work, we select \textbf{AG News}~\cite{zhang2015character}, \textbf{DBPedia}~\cite{lehmann2015dbpedia}, 
\textbf{Yahoo}~\cite{zhang2015character} and
\textbf{NYT}~\cite{nyt} for topic classification, and 
\textbf{IMDB}~\cite{imdb}, 
\textbf{SST-2}~\cite{socher-etal-2013-recursive},
\textbf{Amazon}~\cite{amazon}\footnote{We follow \cite{hu-etal-2022-knowledgeable} to subsample a 40K subset from the original 400K test data for faster evaluations, which has little effect on the average performance in our pilot studies.}, 
\textbf{MR}~\cite{mr}, 
\textbf{Yelp}~\cite{zhang2015character} for sentiment analysis.
All the datasets are in English. 
We report performance on the test set when available, falling back to the validation set for SST-2. 
The details for these datasets can be found in table \ref{tab:dataset}. 

\noindent  $\diamond$ \textbf{Corpus.} We follow~\citep{shi2022nearest,van2022don} to obtain a heterogeneous collection of text that are broadly relevant to tasks in our experiments as the general-domain unlabeled corpus $\cD$. 
Specifically, we select \textsc{Wiki}~\cite{petroni-etal-2021-kilt}, subsets of \textsc{Reviews}~\cite{review} and \textsc{RealNews}~\cite{realnews} to form the corpus. The detailed information and preprocessing steps for these corpora are shown in Appendix~\ref{sec:corpus_apd}.

\noindent  $\diamond$ \textbf{Metrics.} 
We use F1 score as the metric for NYT as the label distribution is imbalanced. Accuracy is used for the remaining tasks.

\noindent $\diamond$ \textbf{Baselines.} We consider various baselines, including both zero-shot inference and dataset generation methods. Details of the baselines are in Appendix~\ref{sec:bsl_apd}.
We also list the results with extra resources (\eg large PLMs, task-specific samples, or knowledge bases), but only for reference purposes, \emph{since we do not claim {\ours} achieves state-of-the-art performance on zero-shot text classification. 
Rather, we consider {\ours} as a better approach to synthesizing datasets in a zero-shot manner for  text classification tasks.} 

\noindent  $\diamond$ \textbf{Implementation Details.} 
For implementation, we use PyTorch~\cite{paszke2019pytorch} and HuggingFace~\cite{wolf2019huggingface}. We set the retrieval rounds $T=3$, the $k$ used in ANN in Eq.~\ref{eq:knn} to $100$ for the 1st round and $20$ for later rounds in Eq.~\ref{eq:knn_expansion}. 
The number of the training data per class is set to no more than 3000~\cite{meng2022generating}. 
Under the zero-shot learning setting, we keep all hyperparameters the \emph{same} across all tasks due to the lack of validation sets.  
In principle, {\ours} is compatible with any dense retriever $R_{\theta}$ and classifier $C_{\phi}$. In this work, we initialize $R_{\theta}$ from  Condenser~\cite{gao-callan-2021-condenser} and fine-tune RoBERTa-base~\cite{liu2019roberta} as $C_{\phi}$. See App.~\ref{sec:impl_apd} for details.

\subsection{Main Experiment Results}
The results of {\ours} and compared baselines on nine tasks are in Table~\ref{tab:main_f1}. From these results, we have the following observations:

\noindent (1) {\ours} significantly surpasses fair baselines on average of nine datasets, and often achieves comparable or even better results against methods using extra task-specific information. Compared with our direct baseline  \cite{van2022don} using regular expressions to mine training data, {\ours} achieves 4.3\% gain on average. The gain is more notable (6.8\%) for topic classification with more classes. These results justify that dense retrieval serves as a more flexible way to understand the category and can extract training data being  \emph{semantically closer} to the target topics.

\noindent (2) SuperGen~\cite{meng2022generating} achieves strong results on sentiment tasks. 
 However, its performance diminishes for multi-class topic classification, suggesting that NLG-based dataset generation methods may struggle to produce sufficiently accurate and distinct texts for fine-grained classification.

\noindent (3) {\ours} also delivers competitive performance against zero-shot learning and weakly-supervised text classification baselines without requiring additional resources, such as larger language models or task-specific unlabeled data. This suggests that dataset generation serves as an alternative approach for zero-shot text classification.

\begin{table}[t]
\vspace{-0.05em}
\centering
\renewcommand\arraystretch{0.9}
\resizebox{\columnwidth}{!}{
\begin{tabular}{l*{4}{c}}
\toprule
{\textbf{Method}} & \textbf{AG News} & \textbf{DBPedia} & \textbf{SST-2} & \textbf{Yelp} \\ 
\midrule
\textbf{\ours} &  \bf 85.0 & \bf 87.6 & \bf 88.9 & \bf 93.0\\
\ w/o Data Curation (DC) & 70.9   &	68.8 & 69.2 & 75.5 \\
\ w/o Multi-step Retrieval (MSR)  & 83.0 & 83.6 & 85.9 & 90.9\\
\ w/o Label Smoothing (LS) & 84.5 &	86.1 & 88.0 & 91.7 \\
\bottomrule
\end{tabular}
}
\caption{
Ablation Study. For w/o DC, we use $R_{\theta}$ to calculate similarity between samples and labels for zero-shot learning. 
For w/o MSR, we only retrieve \emph{the same size of} data as {\ours} for one round with verbalizers. For w/o LS, we use one-hot labels for fine-tuning. 
}
\vspace{-0.5em}
\label{tab:ablation}
\end{table}

\begin{figure}[t]
\centering
\includegraphics[width=0.9\linewidth]{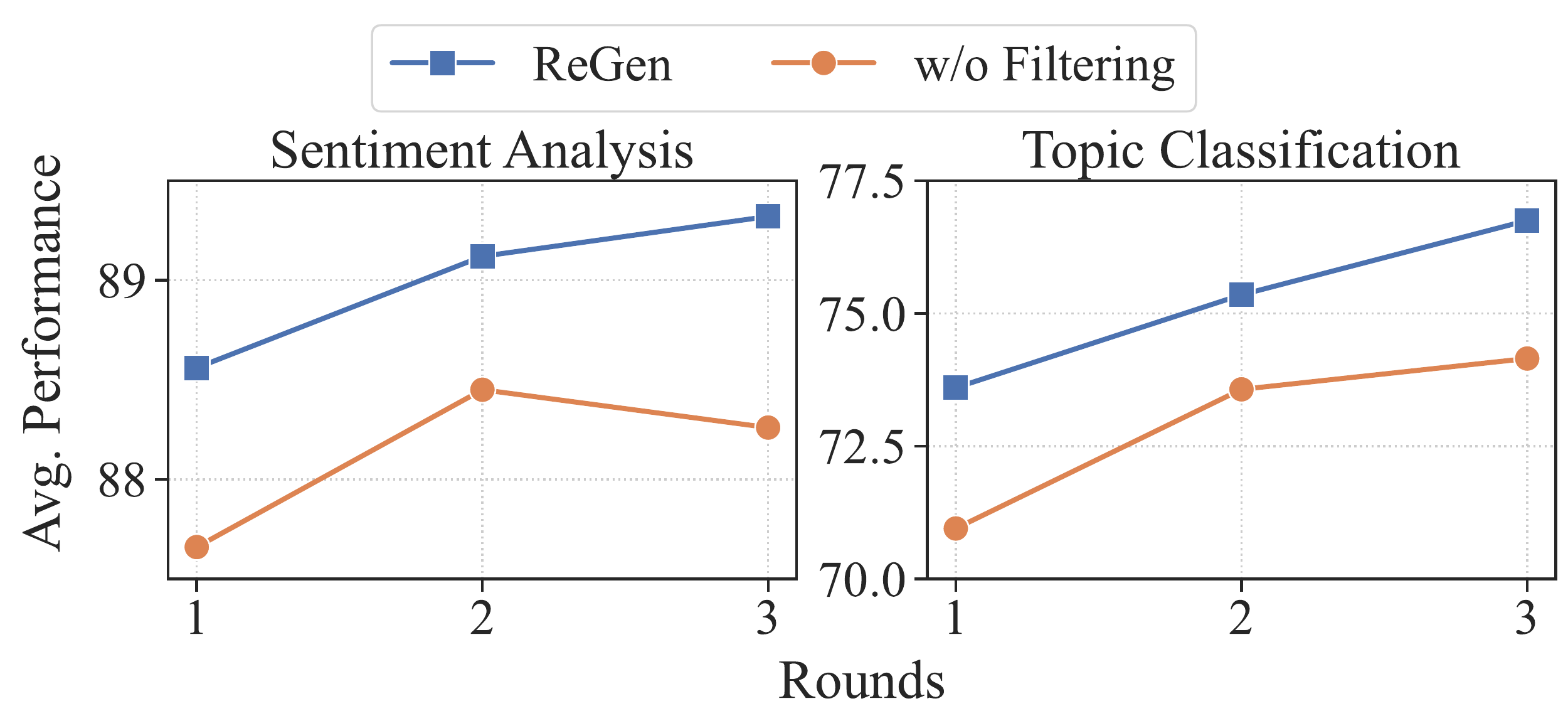}
\vspace{-1.5ex}
\caption{Effect of self-consistency guided filtering. }
\vspace{-2ex}
\label{fig:filtering}
\end{figure}
\subsection{Ablation Studies}

\noindent \textbf{Effect of Different Components.} Table~\ref{tab:ablation} shows the result of ablation studies on four datasets\footnote{More results on other datasets are in Appendix~\ref{sec:per_task_results}.}, which demonstrates the superiority of retrieving texts from the corpus for training data creation as well as conducting multi-step retrieval. Besides, label smoothing also results in  performance gain as it mitigates the effect of noisy labels for fine-tuning. 

Besides, we plot the result over different rounds of retrieval in Fig.~\ref{fig:filtering}. It is clear that both multi-step retrieval and filtering progressively enhance the performance of target tasks, justifying their necessity for improving the quality of training data. \emph{We have also attempted to conduct more retrieval rounds, but do not observe significant performance gains}.


\begin{figure}
	\centering
	\vspace{-1.5ex}
	\subfigure[Topic]{
		\includegraphics[width=0.47\linewidth]{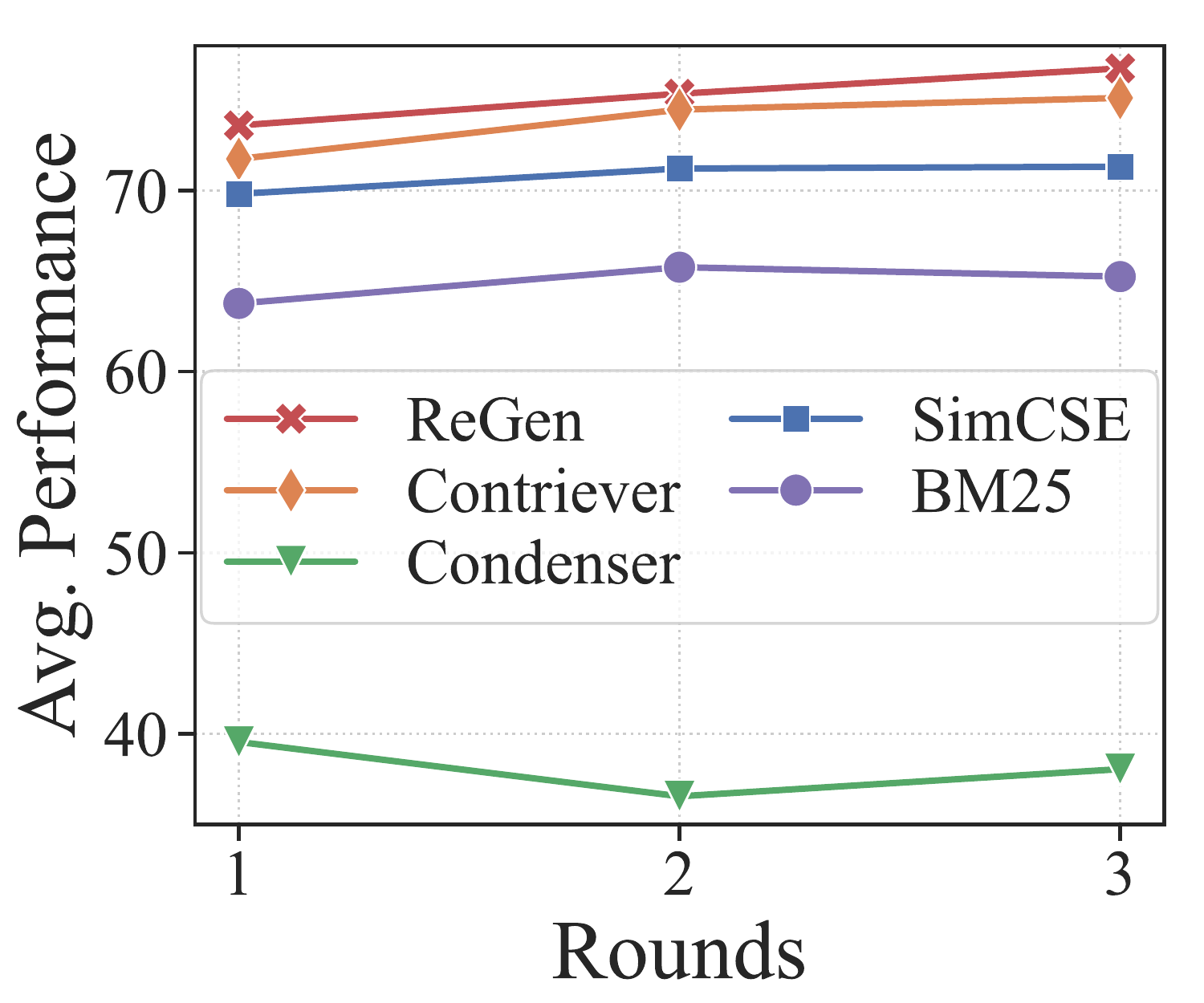}
		\label{fig:encoder_topic}
	} \hspace{-1ex} 
	\subfigure[Sentiment]{
		\includegraphics[width=0.47\linewidth]{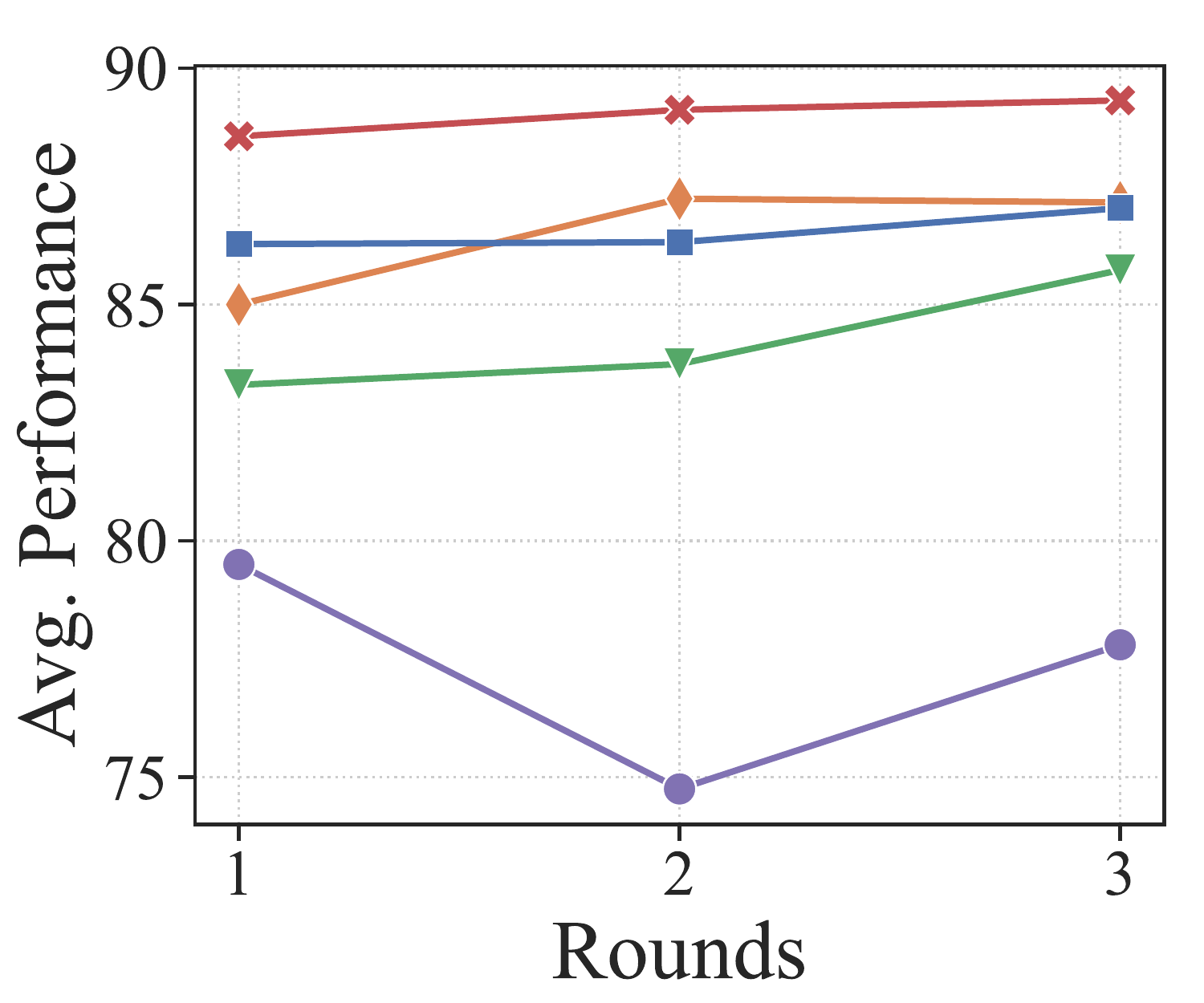}
		\label{fig:c}
	}
	\vspace{-1ex}
	\caption{Effect of different dense retrieval models $R_{\theta}$.}
	\vspace{-2ex}
\label{fig:dr_model}
\end{figure}
\noindent \textbf{Study of Dense Retrievers.} 
We compare the retrieval model $R_{\theta}$ with other off-the-shelf unsupervised retrieval models. Here we choose one sparse model BM25~\cite{bm25} and three DR  models: Condenser~\cite{gao-callan-2021-condenser}, SimCSE~\cite{gao-etal-2021-simcse}, and Contriever~\cite{izacard2021contriever}. 
From Figure~\ref{fig:dr_model}, we observe that the performance of BM25 is not satisfactory, since simply using lexical similarity is insufficient to retrieve a set of diverse documents for fine-tuning. 
Besides, our retrieval model outperforms other unsupervised DR models for two reasons: (1) Condenser and SimCSE are pretrained over \emph{short sentences}, and the learning objective is suboptimal for long documents; (2) these models are not pretrained on the corpus used in our study and suffer from the \emph{distribution shifts}~\cite{yu2022cocodr}. Instead, our strategy can better adapt the PLM for the retrieval task.

In the following sections, we mainly compare {\ours} with Mining~\cite{van2022don} and SuperGen~\cite{meng2022generating} as they are  closest baselines to us.
\begin{figure}[!t]
	\centering
	\vspace{-1ex}
	\subfigure[Topic]{
		\includegraphics[width=0.47\linewidth]{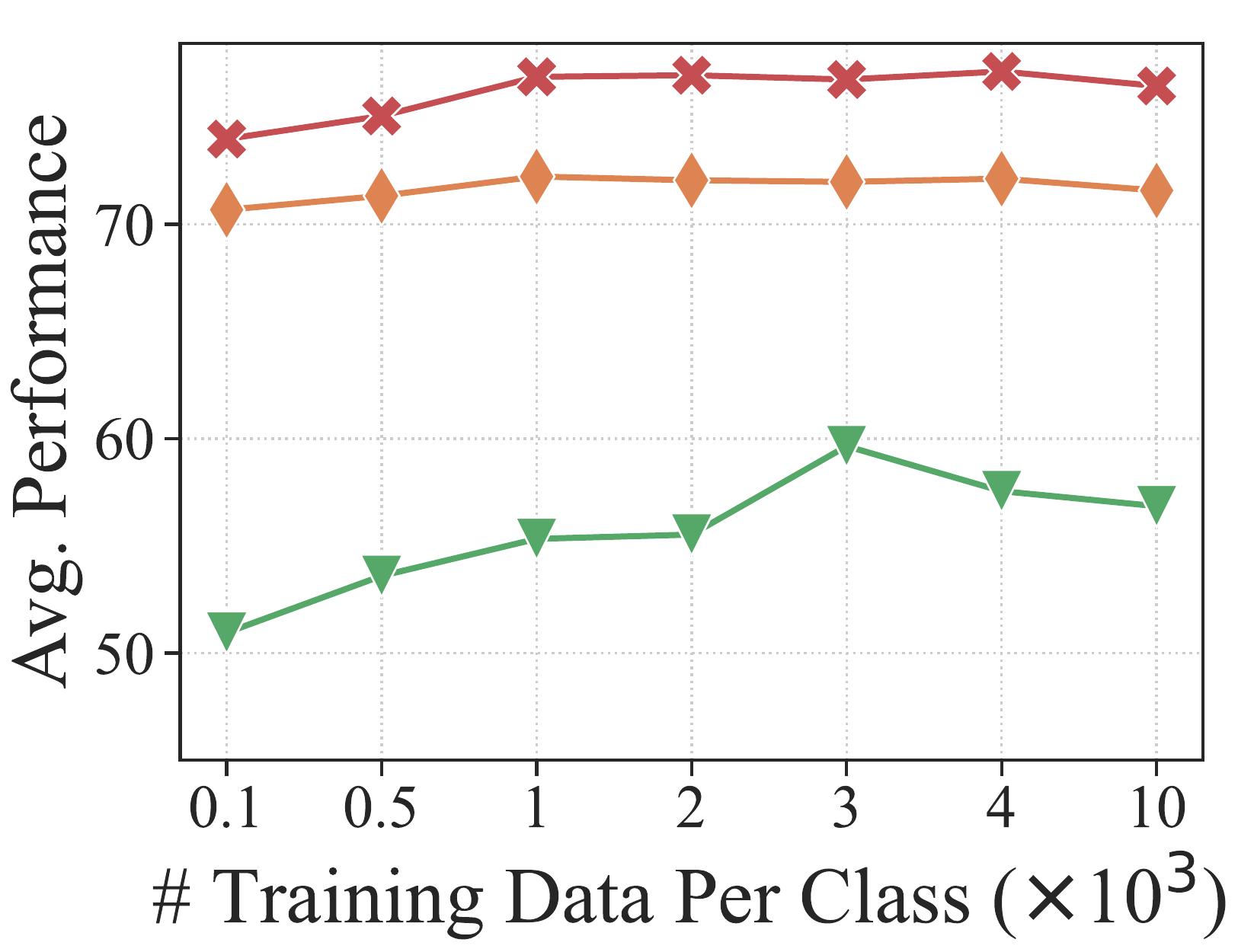}
		\label{fig:encoder_topic}
	} \hspace{-1ex} 
	\subfigure[Sentiment]{
		\includegraphics[width=0.47\linewidth]{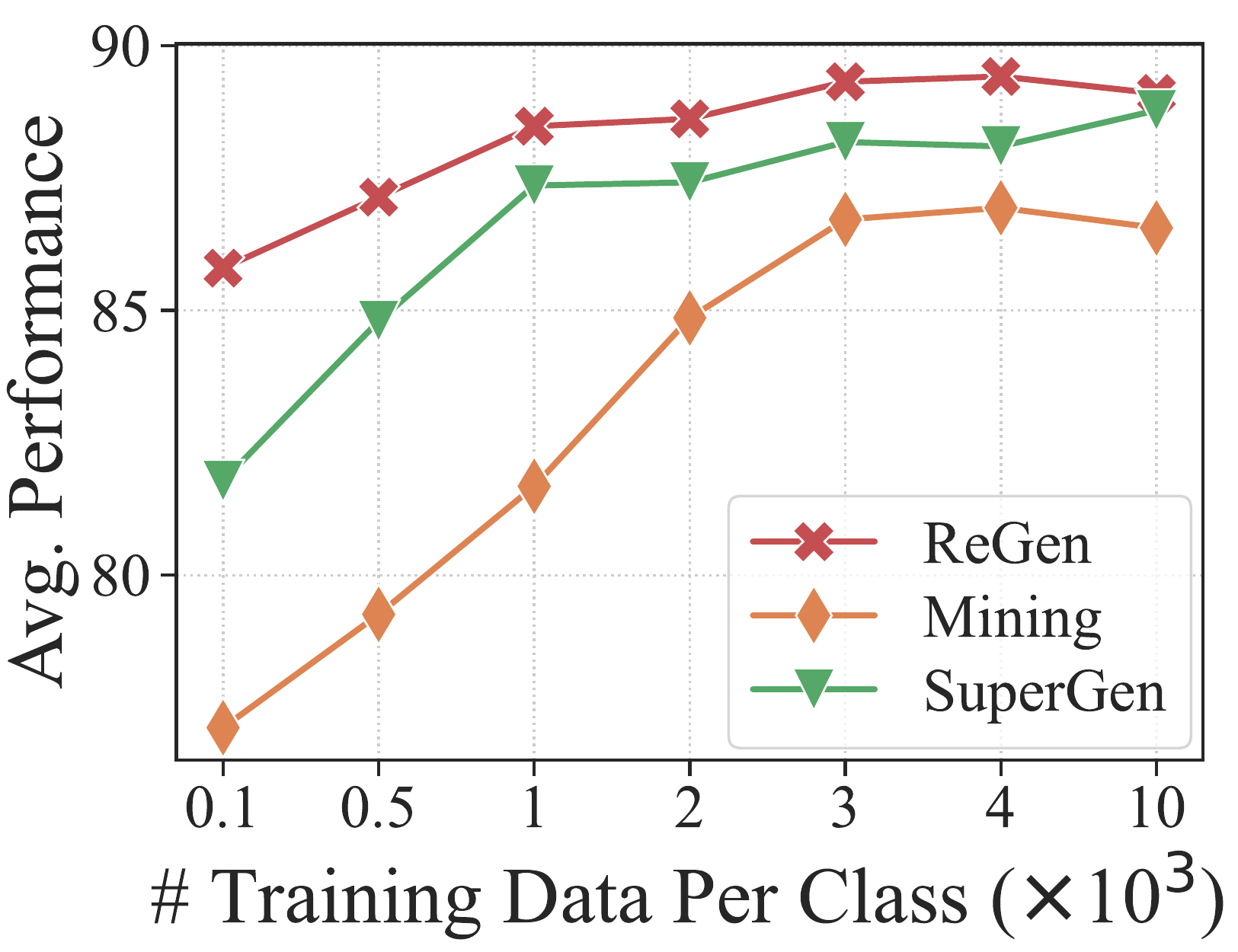}
		\label{fig:c}
	}
	\vspace{-2ex}
	\caption{Effect of the training data size.}
	\vspace{-2ex}
\label{fig:num_train_data}
\end{figure}

\subsection{Effect of the Amount of Generated Data}
Figure~\ref{fig:num_train_data} shows the results of using different amount of training data (after filtering). 
Overall, we find that the performance first improves significantly when the number of training data is small (\eg, $100$), then becomes stable with more retrieved data. This is because with too many generated data, it may also introduce more label noise and reduce the quality of training data. 
Nevertheless, {\ours} outperforms baselines under different volumes of training samples, justifying its advantage.

\begin{figure}[!t]
	\centering
	\vspace{-1ex}
	\subfigure[AG News]{
		\includegraphics[width=0.47\linewidth]{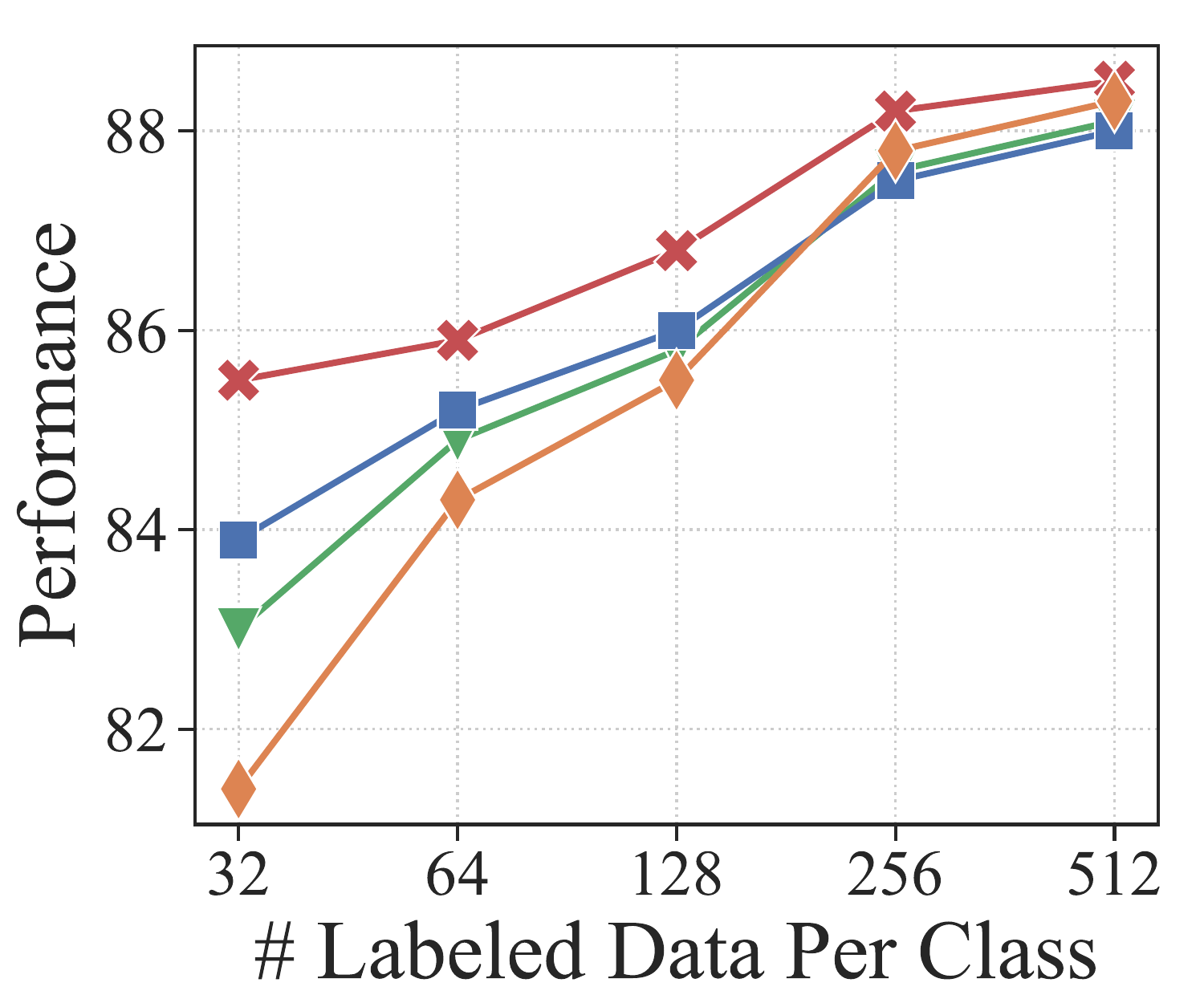}
		\label{fig:encoder_topic}
	} \hspace{-1ex} 
	\subfigure[IMDB]{
		\includegraphics[width=0.47\linewidth]{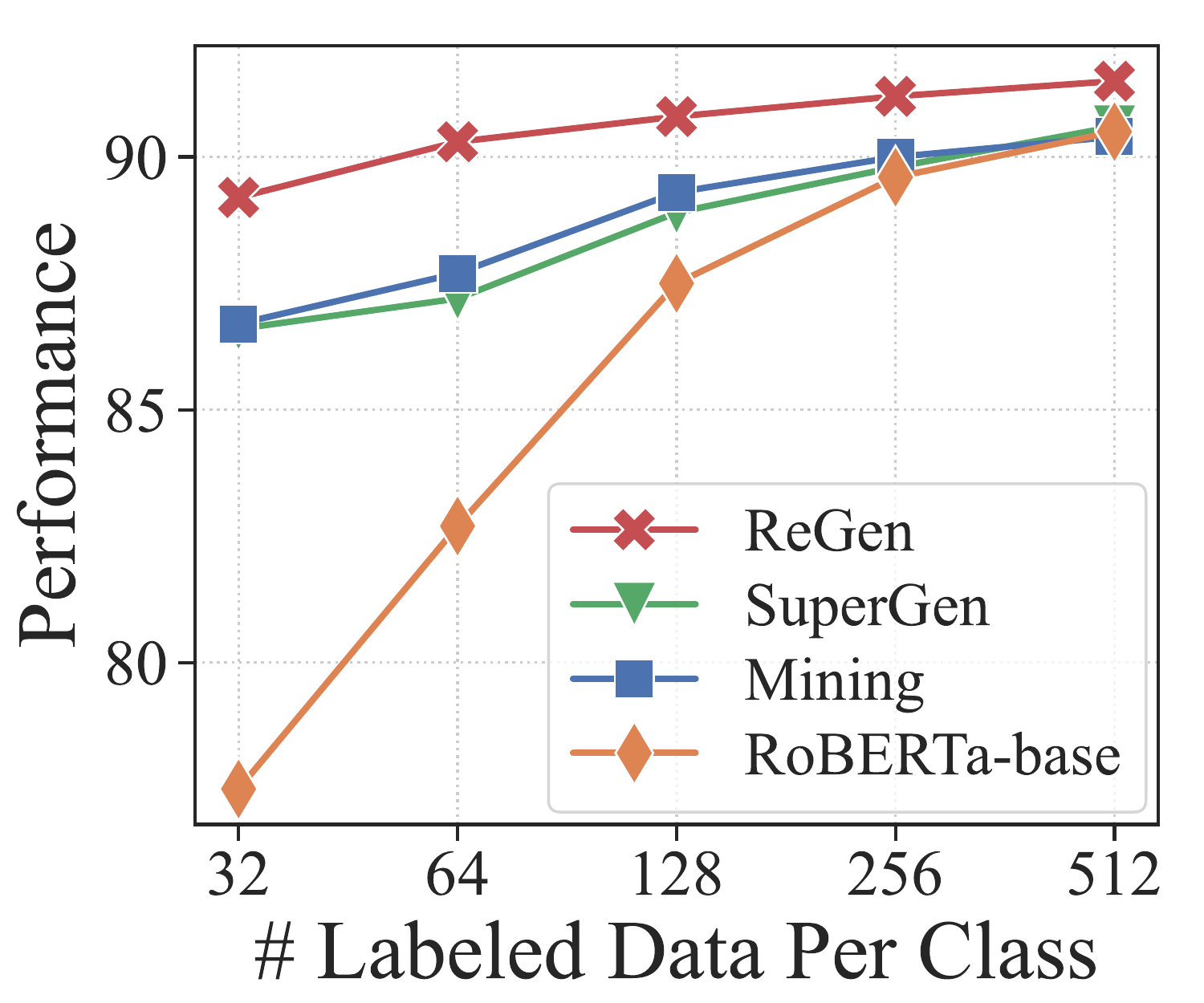}
		\label{fig:c}
	}
	\vspace{-1.5ex}
	\caption{Accuracy on IMDB/AG News fine-tuned on the few labeled samples only vs. on the few-shot and generated dataset with varying amount of labeled data.}
	\vspace{-2ex}
\label{fig:few_shot}
\end{figure}
\begin{table}[t]
\vspace{-0.1em}
\centering
\renewcommand\arraystretch{0.9}
\resizebox{\columnwidth}{!}{
\begin{tabular}{l*{4}{c}}
\toprule
\multirow{2}{*}{\textbf{Dataset}} & \multicolumn{2}{c}{\textbf{AG News}} & \multicolumn{2}{c}{\textbf{DBPedia}} \\ 
& {\ours} & {\ours}+LLM & {\ours} & {\ours}+LLM \\
\midrule
\bf Accuracy & 85.0$\pm${0.8} & 85.4$\pm${0.5} & 87.6$\pm${0.9} & 88.5$\pm${0.8} \\
\bottomrule
\end{tabular}
}
\vspace{-1ex}
\caption{
Effect of using Large Language Models for Verbalizer Expansion.
}
\vspace{-1ex}
\label{tab:llm_aug}
\end{table}

\subsection{Fusing {\ours} with Large Language Models (LLMs)}
In this section, we give a simple demonstration of how to leverage recently-proposed large language models (e.g. GPT-4~\cite{openai2023gpt4}) to further boost the performance. 
As LLMs have demonstrated strong ability for text generation, we use them to augment the verbalizer before retrieving documents from the general-domain corpus. The details are in Appendix \ref{sec:llm_expan}.

From Table \ref{tab:llm_aug}, we observe that expanded verbalizers lead to consistent performance gains on two datasets. 
Although the scale of the improvement is not that significant, it shows some effectiveness with such cheap plug-in techniques of using LLMs for boosting {\ours}.

\subsection{Using {\ours} in Few-Shot Settings}
{\ours} can also be combined with a few labeled examples to improve the performance. 
We follow \cite{meng2022generating} to fine-tune $C_{\phi}$ with few-shot examples and the synthetic dataset (Details in Appendix \ref{sec:few_apd}) using IMDB and AG News as examples. From Fig.~\ref{fig:few_shot}, we observe that {\ours}  improves over the vanilla few-shot fine-tuning under all studied regions ($32$ to $512$ labels per class), while baselines cannot further promote the performance with more training samples. Quantitatively, the performance of {\ours} is equivalent to that of fine-tuning with 128-256 labeled documents per class. With 32 labels per class, {\ours} achieves comparable performance of vanilla fine-tuning with 4x-8x labeled examples. These results verify that {\ours} promotes label efficiency of PLMs. 

\subsection{Robustness over Different Verbalizers}
\label{sec:verbalizers}
\begin{table}[t]
\centering
\renewcommand\arraystretch{0.9}
\resizebox{\columnwidth}{!}{
\begin{tabular}{l*{4}{c}}
\toprule
\bf Dataset & {\textbf{Verbalizer Group}} & \bf Mining & \bf SuperGen & \bf{\ours} \\ 
\midrule
\multirow{5}{*}{IMDB} &\# 0 (Original) & 87.1 &	85.8& \bf	89.9\\
&\# 2  & \bf 88.3&	82.7& 	87.9 \\
&\# 3  &86.0 &	80.6 & \bf	90.3 \\
&\# 4  &89.0	&89.1	&\bf 90.1 \\
\cmidrule(l){2-5} 
& Avg. $\pm$ Std. & 87.6$\pm${1.3} & 84.5$\pm${3.6} & \bf 89.6$\pm${1.2}\\
\midrule
\multirow{5}{*}{AG News} &\# 0 (Original) & 79.7 &	77.4 &	\bf 85.0 \\
&\# 1 & 82.5 &	75.2 & 	\bf 82.7 \\
&\# 2 & 83.9 &	77.8 &	\bf 85.1 \\
&\# 3 & 77.7 &	72.2 &	\bf 83.6 \\
\cmidrule(l){2-5} 
& Avg. $\pm$ Std. &  80.9$\pm$2.7 & 75.6$\pm$2.5 & \bf 84.2$\pm$1.2 \\
\bottomrule
\end{tabular}
}
\caption{
Results with different verbalizers. The number for each prompt group is the averaged performance across 5 runs; Avg.$\pm$Std. is calculated over four groups.
}
\vspace{-1ex}
\label{tab:prompt_study}
\end{table}

As {\ours} and zero-shot dataset generation methods always rely on a class-dependent verbalizer to steer the whole process, we study the impact of different verbalizers on the final performance. We use IMDB and AG News as two datasets, and create three different groups of verbalizers other than the default ones for comparison (Details in Appendix \ref{sec:prompt_apd}).
From Table \ref{tab:prompt_study}, we observe that {\ours} generally outperforms baselines on 7 out of 8 cases. {\ours} also has \emph{lower} performance variance across four groups of verbalizers.  These results reveal that {\ours} does not rely on specific designs of verbalizers, and are more robust over different verbalizers.

\begin{figure}[!t]
	\centering
	\vspace{-1.5ex}
	\subfigure[]{
		\includegraphics[width=0.46\linewidth]{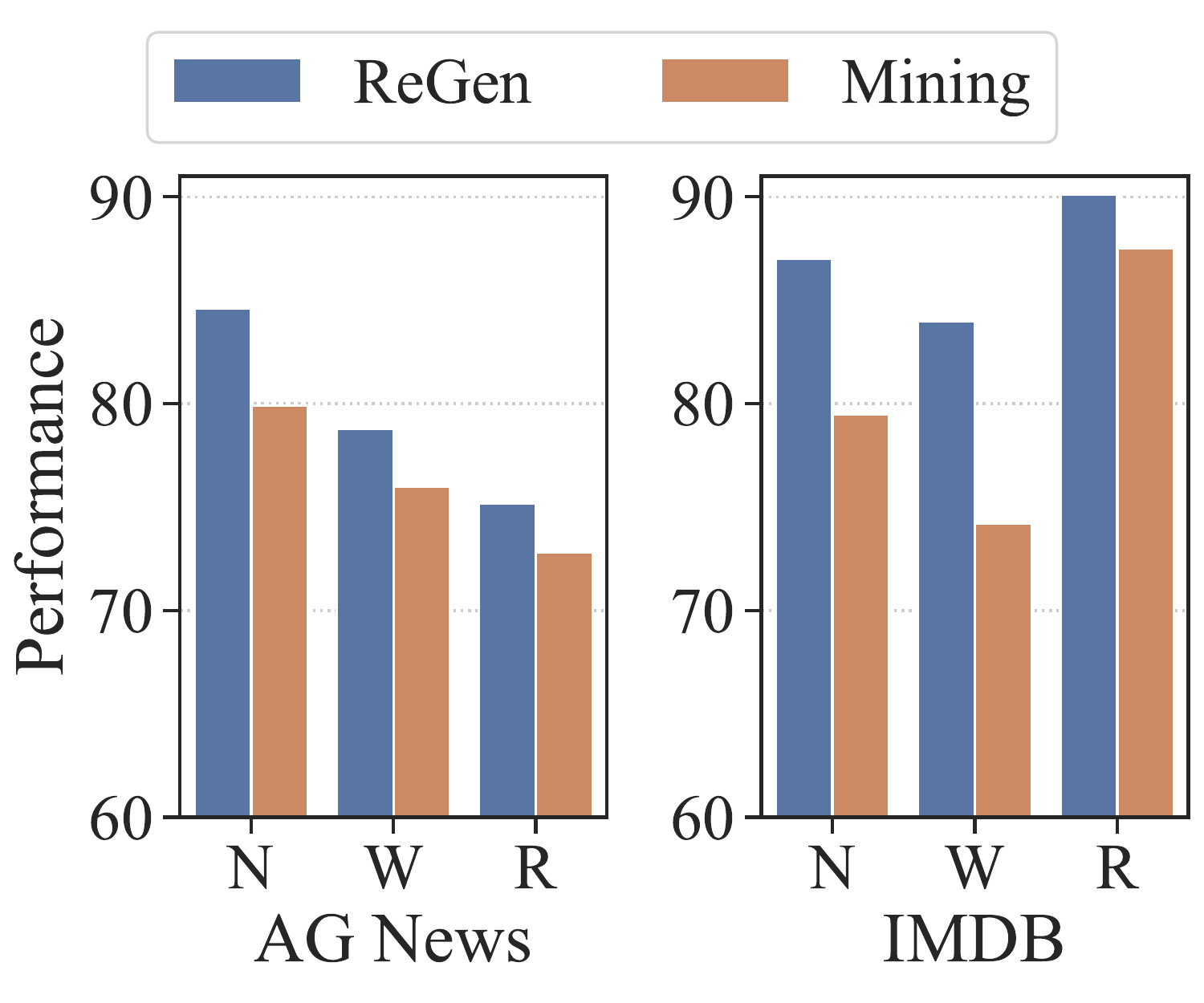}
		\label{fig:subset}
	} 
	\subfigure[]{
		\includegraphics[width=0.46\linewidth]{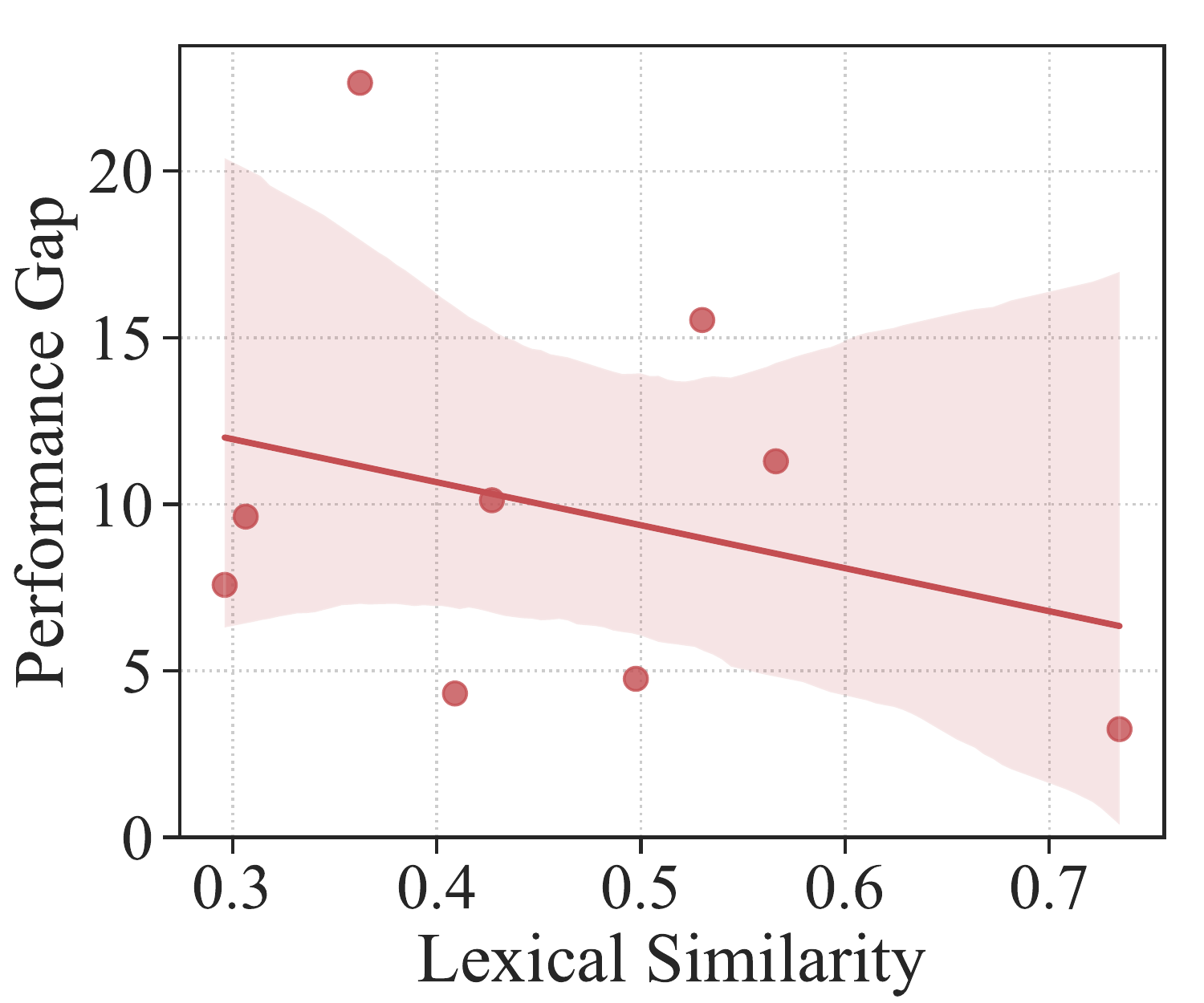}
		\label{fig:gap_similarity}
	}
	\vspace{-2.5ex}
	\caption{(a) Performance of {\ours} and Mining using only subset of corpus $\cD$. N/W/R stands for \textsc{RealNews}/\textsc{Wiki}/\textsc{Reviews}, respectively. (b) The relation on the performance gap and lexical similarity between the corpus and target tasks.}
	\vspace{-2ex}
\label{fig:corpus}
\end{figure}

\subsection{The Effect on General-domain Corpus $\cD$}
\label{sec:corpus}
We study the effect of corpus $\cD$ by conducting retrieval on different subsets from $\cD$. 
As shown in Figure~\ref{fig:subset}, we observe better performance when the corpus aligns with the target task well (\eg \textsc{News} for AG News). This is expected as the model suffers less from the distribution shift issue. 
Besides, {\ours} outperforms the mining method under all settings, justifying its superior ability to retrieve relevant text even if there is a domain mismatch between the task and corpus. 

Fig.~\ref{fig:gap_similarity} exhibits the relation on the lexical similarity (measured by weighted Jaccard score), and the performance gap between {\ours} and fully-supervised BERT (details in Appendix~\ref{sec:jaccard_apd}).  
Overall, there is a negative correlation among performance gaps and the distribution similarities, as {\ours} performs closer to fully-supervised models on tasks where task-specific documents share more similar lexical patterns with the general-domain corpus. 

\subsection{Efficiency Studies}\label{sec:efficiency}
\begin{table}[t]
\vspace{-0.1em}
\centering
\renewcommand\arraystretch{0.9}
\resizebox{\columnwidth}{!}{
\begin{tabular}{l*{3}{c}}
\toprule
{\textbf{Operation}} & \textbf{Mining} & \textbf{Supergen} & \textbf{\ours} \\ 
\midrule
{Pretraining} & --- & --- & 23h \\
Indexing of Corpus/Per doc & ---  &	--- & 6h/4ms \\ \midrule
{Curating Dataset Per Task} & 1.4h & 20.4h & 0.6h \\
Filtering Per Task  & 0.2h &	0.1h & 0.5h \\
Model Fine-tuning Per Task & 0.4h &	0.3h & 0.7h  \\\midrule
\bf Total Time (for all Tasks) & 10h & 104h & 38h \\
\bottomrule
\end{tabular}
}
\vspace{-1ex}
\caption{
Efficiency Study. For {\ours}, the average time per task of curating dataset, filtering and fine-tuning is accumulated over 3 rounds. 
}
\vspace{-1.5ex}
\label{tab:efficiency}
\end{table}

Table~\ref{tab:efficiency} measures the efficiency of {\ours} and baselines. While the pretraining and indexing corpus for {\ours} can be time-consuming, it only needs to be done once, thus the overall running time of {\ours} is significant lower than the baseline using large NLG models~\cite{meng2022generating}. Compare with the mining-based method, although {\ours} costs longer time in total, we think it is worthwhile as {\ours} outperforms it on all nine tasks studied in this work.

\subsection{Quality Analysis of Synthetic Datasets}
\label{sec:quality}
We provide other measurements to better evaluate the quality of the generated dataset of {\ours} and baselines~\cite{ye2022progen}. 

\noindent \textbf{Automatic Evaluations.}
\begin{table}[t]
\centering
\renewcommand\arraystretch{0.9}
\resizebox{\columnwidth}{!}{
\begin{tabular}{l*{4}{c}}
\toprule
\bf Dataset & {\textbf{Metrics}} & \bf Mining & \bf SuperGen & \bf{\ours} \\ 
\midrule
\multirow{3}{*}{Sentiment} & Correctness $(\uparrow)$ & 0.815 &	0.971 & \bf	0.986 \\
& Diversity $(\downarrow)$ & \bf 0.144 &	0.915 &	0.361 \\
& Distribution Sim. $(\uparrow)$ &0.856	 & 0.803 & \bf	0.865\\
\midrule
\multirow{3}{*}{Topic} & Correctness $(\uparrow)$ & 0.759 &	0.626 &	\bf 0.860 \\
& Diversity $(\downarrow)$ & \bf 0.132	 & 0.767 &	0.346 \\
& Distribution Sim. $(\uparrow)$  & 0.748 &	0.648 & \bf	0.757 \\

\bottomrule
\end{tabular}
}
\vspace{-1ex}
\caption{
Automatic evaluation results on three metrics.
}
\vspace{-1.5ex}
\label{tab:auto_eval}
\end{table}

\begin{table}[t]
\centering
\renewcommand\arraystretch{0.9}
\resizebox{\columnwidth}{!}{
\begin{tabular}{l*{4}{c}}
\toprule
\bf Dataset & {\textbf{Metrics}} & \bf Mining & \bf SuperGen & \bf{\ours} \\ 
\midrule
\multirow{3}{*}{Sentiment} & Correctness $(\uparrow)$ & 1.46 & \bf 1.95 & 1.94\\
& Diversity $(\uparrow)$& \bf 2.00 & 0.75 &\bf  2.00\\
& Informativeness $(\uparrow)$&1.40 & 1.90  &\bf  1.92\\
\midrule
\multirow{3}{*}{AG News} & Correctness $(\uparrow)$& 1.78 &1.74  & \bf 1.94\\
& Diversity $(\uparrow)$& 1.62 &  0.94 & \bf 1.88 \\
& Informativeness $(\uparrow)$& 1.63 & 1.43 &\bf 	1.82\\
\bottomrule
\end{tabular}
}
\vspace{-1ex}
\caption{
Human evaluation results on three metrics. (The full score is 2)
}
\vspace{-1em}
\label{tab:human_eval}
\end{table}

We first measure the quality of the dataset from three perspectives: \emph{correctness}, \emph{diversity}, \emph{distribution similarity}. The details are shown in Appendix~\ref{sec:autoeval_apd}.
Overall, the diversity of generated text from NLG models~\cite{meng2022generating} is not satisfactory, and the correctness of text from NLG models is also not guaranteed for topic classification tasks. 
For the mining-based method, despite it achieves better diversity, the performances on other two metrics are worse. As a result, {\ours} surpasses it on these tasks.

\noindent \textbf{Human Evaluations.}
We also conduct human evaluations to evaluate the quality of the synthetic dataset using AG News and Sentiment datasets as two examples. For each class, we randomly sample 25 documents  and ask 4 human volunteers to evaluate the dataset from three perspectives: \emph{Correctness}, \emph{Informativeness} and \emph{Diversity} (details in Appendix~\ref{sec:human_eval_apd}).  
The mean ratings are shown in Table~\ref{tab:human_eval}. 
The average Fleiss' Kappa~\cite{fleiss1971measuring} for correctness, informativeness and diversity are 0.53/0.57/0.58 (Moderate Agreement), respectively.
Overall, the dataset curated by {\ours} has the best informativeness and diversity, while has a competitive result on correctness score. 
These results indicate that {\ours} improves over previous works for curating a better dataset to tackle the downstream tasks. 
Detail cases of samples from the synthetic datasets can be found at Appendix~\ref{sec:case_study}.

\section{Discussion and Conclusion}
\vspace{-0.5ex}
\subsection{Discussion}
\paragraph{Extending {\ours} to Specific Domains.} The {\ours} framework is versatile and can be applied to various domains beyond our experiments. For example, it is possible to extend {\ours} to zero-shot biomedical text classification~\cite{scidocs} using the  publicly available PubMed articles as the unlabeled corpus.  

\paragraph{Verbalizers Selection for {\ours}.} All the  verbalizers used in this work are from the prior works~\cite{hu-etal-2022-knowledgeable,schick-schutze-2021-exploiting} to circumvent manual prompt engineering and ensure a fair comparison. For those datasets where verbalizers are not given, we can adopt automatic verbalizer and template generation approaches~\cite{gao-etal-2021-making} to generate verbalizers for retrieving relevant documents. 

\paragraph{Soliciting Human Feedbacks to Improve {\ours}.} 
In many cases, there may exist difficult examples where the classifier and the retrieval model do not agree with each other. To enable the model to learn on these hard examples, \emph{active learning} can be adopted to solicit human annotations~\cite{yuan-etal-2020-cold,yu-etal-2022-actune,yu2022cold} or instructions~\cite{peng2023check,zhang-etal-2022-prompt,zhang2022adaptive} to further improve the model performance.

\paragraph{Collaboration with Large Language Models.} There are many other potential ways to incorporate black-box large language models into {\ours} beyond our experiments. 
For instance, large language models can be used to \emph{rerank} the top retrieved documents~\cite{ma2023zero} or generate \emph{augmented examples} for classifiers~\cite{moller2023prompt}.
On the other hand, {\ours} can be integrated into the training set synthesis for language models when the labeled dataset is inaccessible~\cite{zhang2023blindly}.
It is still an open question on how to harness large language models for dataset generation in an  efficient and effective way. 
\subsection{Conclusion}
In this paper, we propose a framework {\ours} for zero-shot text classification, which incorporates dense retrieval to synthesize  task-specific training sets via retrieving class-relevant documents from the generic unlabeled corpus with verbalizers. We further propose two simple while effective strategies to progressively improve the quality of the curated dataset.  
The effectiveness of {\ours} is validated on nine benchmark datasets with an average gain of 4.3\%. Further qualitative analysis justify the better quality of datasets generated by {\ours} over baselines under multiple criteria. 

\section*{Limitations}
Our method {\ours} is a general framework for zero-shot text classification. In this work, we aim to first bring in simple and intuitive way to justify the power of unsupervised dense retrieval for zero-shot learning. 
Effective as it is, there is still much room for improvements, including designing better objectives for pretraining $R_{\theta}$ as well as better strategies for removing noisy training data~\cite{lang2022training,xu2023neighborhood}. How to improve these components is an important line of future work.

Besides, our experiment results are all based on BERT$_{\text{base}}$ sized models. Although {\ours} performs on par with or better than previous dataset generation methods using giant NLG models, it remains unknown to us how the benefit of {\ours} scales with more parameters for both $R_{\theta}$ and $C_{\phi}$.  

Also, we point out that this work focuses on zero-shot \emph{text classification} with task-specific verbalizers and unlabeled generic corpus, thus it can be nontrivial  to adapt our framework to other tasks such as Natural Language Inference (NLI) as well as low-resource tasks where even the unlabeled generic corpus can be hard to collect. 
Extending {\ours} to these settings will reduce the annotation burden under more challenging scenarios.

\section*{Ethics Statement}
\vspace{-0.5em}
One potential risk of applying {\ours} is that the generic corpus used in our experiments may contain harmful information as they were crawled from the Internet that are only filtered with some  rules~\cite{gehman-etal-2020-realtoxicityprompts}. As a result, they may contain  text exhibiting biases that are undesirable for target tasks. 
To alleviate this issue, we recommend the potential users to first use bias reduction and correction techniques~\cite{schick2021self} to remove biased text from the corpus to mitigate the risks of the curated dataset.

\section*{Acknowledgements}
\vspace{-0.5em}
We thank anonymous reviewers for their feedback. 
This work was supported by NSF IIS-2008334, IIS-2106961, and CAREER IIS-2144338.

\bibliography{anthology,custom,emnlp22,datasets}
\bibliographystyle{acl_natbib}

\clearpage
\appendix

\section{Verbalizers and Templates for Datasets}
\label{sec:dataset_apd}
The verbalizers and  templates of datasets are shown in table \ref{tab:prompts}.

\begin{table*}[t]
\centering
\resizebox{1.9\columnwidth}{!}{
\begin{tabular}{llll}
\toprule
\textbf{Task} & \textbf{Verbalizers} & \textbf{Template used for Retrieval}  & \textbf{Template used for Prompting} \\
\midrule
\multirow{2}{*}{\textbf{AG News}} & politics, sports & \multirow{2}{*}{\texttt{[VERB]} News.} & \multirow{2}{*}{The category of  $\boldsymbol{x}^b$ is \texttt{[VERB]}.}\\
& business, technology & \\
\midrule
\multirow{4}{*}{\textbf{DBPedia}} & company, school, artist, athlete & \multirow{4}{*}{\texttt{[VERB]}} & \multirow{4}{*}{$\boldsymbol{x}^a$ $\boldsymbol{x}^b$? The category of $\boldsymbol{x}^a$ is \texttt{[VERB]}.} \\
& politics, transportation, building, 
&  \\
& river/mountain/lake, village, animal,  \\
&  plant, album, film, book\\
\midrule
\multirow{3}{*}{\textbf{Yahoo}} & society, science, health, school & \multirow{3}{*}{\texttt{[VERB]}} & \multirow{3}{*}{$\boldsymbol{x}^a$ $\boldsymbol{x}^b$? The category of $\boldsymbol{x}^a$ is \texttt{[VERB]}.} \\
& computer, sports, business, 
&  \\
& music, family, politics  \\
\midrule
\multirow{3}{*}{\textbf{NYT}} & business, politics, sports & \multirow{3}{*}{ \texttt{[VERB]} News.} & \multirow{3}{*}{The category of  $\boldsymbol{x}^b$ is \texttt{[VERB]}.} \\
& health, education, estate \\
& art, science, technology \\
\midrule
\multirow{2}{*}{\textbf{Sentiment}} & great & \multirow{2}{*}{It was  a \texttt{[VERB]} movie.} & \multirow{2}{*}{It was a \texttt{[VERB] movie.}. $\boldsymbol{x}^b$.}\\
& bad & \\
\bottomrule
\end{tabular}
}
\caption{
The format of verbalizers and the template used for retrieval and prompting. We use the prompt formats provided in prior works~\cite{schick-schutze-2021-exploiting,hu-etal-2022-knowledgeable}. The \texttt{[VERB]} stands for the verbalizers. 
$\boldsymbol{x}^a$ stands for the title  (only exist in DBPedia and Yahoo) and $\boldsymbol{x}^b$ stands for the body of the target document. 
}
\vspace{-0.5ex}
\label{tab:prompts}
\end{table*}

\section{Corpus}
\label{sec:corpus_apd}
We select three types of corpus, i.e.  \textsc{Wiki}~\cite{petroni-etal-2021-kilt}, subsets of \textsc{Reviews}~\cite{review} and \textsc{RealNews}~\cite{realnews} to form the corpus $\cD$. 
We manually remove documents less than 10 words as we observe that these documents do not contain informative content. 
The detailed information is shown in table~\ref{tab:corpus}.
\begin{table*}[t]
\centering
\resizebox{1.4\columnwidth}{!}{
\begin{tabular}{l*{2}{c}}
\toprule
\bf Corpus & \bf Size & \bf Size after Pre-processing  \\ 
\midrule
Wiki~\cite{petroni-etal-2021-kilt} & 6M & 6M \\
News~\cite{realnews} & 11.9M & 6M \\ 
Reviews~\cite{review} & 24.0M & 4M\\
\bottomrule
\end{tabular}
}
\caption{
The information about the general corpus $\cD$ used in this study.
}
\vspace{-1ex}
\label{tab:corpus}
\end{table*}

\section{Baselines}
\label{sec:bsl_apd}
We consider multiple baselines for zero-shot text classification. The details of these baselines are described as follows. We use $*$ to denote baselines with extra resources or large language models.

\paragraph{Zero-shot Inference Methods}
These methods directly inference over the test set for prediction.

\begin{itemize}[leftmargin=*]
\item \textbf{NSP-BERT}~\cite{sun2021nsp}: It uses the next sentence prediction (NSP) task to perform zero-shot learning. Specifically, it construct prompts for each labels, and use the PLM with the NSP head as the indicator.  
\item \textbf{Prompt}~\cite{schick-schutze-2021-exploiting}: It uses the original masked language modeling (MLM) objective with category-specific verbalizers to infer the true label of each sentence. 
\item \textbf{KNN-Prompt}~\cite{shi2022nearest}: It improves zero-shot prompting by retrieving relevant information from an additional heterogeneous corpus, which achieves better coverage of the verbalizers.
\item \textbf{KPT}$^*$~\cite{hu-etal-2022-knowledgeable}: It uses additional knowledge bases (\eg WordNet) to expand the label word space for verbalizers, for improving prompt-based learning.
\item \textbf{GPT-3}$^*$~\cite{gpt3}: It adopts GPT-3 for zero-shot learning. We use the contextual calibration \cite{zhao2021calibrate} by default as it can improve the zero-shot prediction accuracy.
\end{itemize}

\paragraph{Transfer-learning based Inference Methods}
\begin{itemize}[leftmargin=*]
\item \textbf{TE-NLI*}~\cite{yin-etal-2019-benchmarking}: It uses the model fine-tuned on NLI tasks to perform zero-shot classification. 
\item \textbf{NLI-ST*}~\cite{gera2022zero}: It uses self-training to finetune the model on additional unlabeled task-specific corpus. 

\end{itemize}
We are aware that there exist some other models for generic zero-shot learning on NLP such as FLAN~\cite{flan} and T0~\cite{t0}, we do not compare with them since they leverage the labeled data from some of the datasets evaluated in this work (e.g. AGNews, IMDB, according to their original  paper). It is thus inappropriate to use them under the true zero-shot learning setting, since such models can have unfair advantages due to access to related data during pre-training. 

\paragraph{Weakly-supervised Learning Methods}
This line of methods is close to the general zero-shot learning in the sense that it does not rely on any labeled examples for classification~\cite{shen-etal-2021-taxoclass,bond,zhang2021wrench}. Instead, it 
leverages class-specific verbalizers as well as \emph{task-specific} unlabeled data as \emph{weak supervision} for classification. 

\begin{itemize}[leftmargin=*]
\item \textbf{LOTClass*}~\cite{meng-etal-2020-text}: It first matches the label name with the corpus to find category-indicative words, then trains the model to predict their implied categories with self-training.
\item \textbf{X-Class*}~\cite{wang-etal-2021-x}: It  estimates class representations by adding the most similar word to each class, then obtains the document representation with weighted average of word representations. Finally, the most confidence words are selected to fine-tune the classifier. 
\end{itemize}

Note that we present the results for the two methods, but mainly for \emph{reference} purposes as the setting between these approaches and our work is different. 
\begin{table*}[t]
\centering
\resizebox{2\columnwidth}{!}{
\begin{tabular}{*{11}{c}}
\toprule
$\operatorname{lr}_{\text{ft}}$ &$\operatorname{lr}_{\text{cl}}$ & $\operatorname{bsz}_{\text{ft}}$ & $\operatorname{bsz}_{\text{cl}}$ & $|\tilde{\mathcal{T}}^{T}|$ & $E_1$& $E_2$ & $T$   & $\alpha$ & $\tau$ & $(k_1, k_2, k_3)$ \\
\midrule
1e-5 & 1e-4 &32  & 400 & 3,000  & 5 & 5 & 3 & 0.1 & 1 & $(100, 20, 20)$ for sentiment and $(50, 10, 10)$ for topics  \\
\bottomrule
\end{tabular}
}
\caption{
Hyperparameters on different tasks (they are kept same for all tasks).
$\operatorname{lr}_{\text{ft}}$: Learning rate for fine-tuning;
$\operatorname{lr}_{\text{cl}}$: Learning rate for unsupervised contrastive learning; $\operatorname{bsz}_{\text{ft}}$: Batch size; $\operatorname{bsz}_{\text{cl}}$: Batch size for unsupervised contrastive learning; $|\tilde{\mathcal{T}}^{T}|$: Maximum number of selected training data per class after the final retrieval round; $E_1$: Number of  epochs for fine-tuning; $E_2$: Number of epochs for 
contrastive learning; $T$: Number of retrieval rounds, $\alpha$: Parameter for label smoothing ; $\tau$: Temperature parameter for contrastive learning; $(k_1, k_2, k_3)$:  Parameter $k$ used in ANN in each round.
}
\label{tab:hyperpara}
\end{table*}

\paragraph{Dataset Generation Methods}
These methods generates specific datasets for zero-shot learning. Note that  we use the same pretrained RoBERTa-base model as the classifier and use the same label smoothing loss for fine-tuning.

\begin{itemize}[leftmargin=*]
\item \textbf{SuperGen}~\cite{meng2022generating}: It is one of the representative methods for using large natural language generation models  (NLG) for zero-shot learning. It first uses the NLG model to generate training data with prompts, then selects data with highest generation probability for fine-tuning.
\item \textbf{Mining}~\cite{van2022don}: It uses regular expressions with category-related keywords to mine samples (the \emph{next} sentences of the matched text) from the corpus to generate training data. Then, it uses the zero-shot prompting to filter the noisy sample and fine-tune another classification model on the filtered dataset. For fair comparison, we use the same corpus $\cD$, prompt format as ours for zero-shot learning, note that these often result in better performance.
\end{itemize}
The comparison of {\ours} with other methods within this category (\eg \cite{ye2022progen,ye2022zerogen}) is shown in Appendix~\ref{sec:compare_zerogen}. 

\section{Implementation Details}
\label{sec:impl_apd}
\subsection{Implementation Details for Baselines}
For \emph{zero-shot inference} methods, we directly use the numbers from the original papers if available, and reimplement Dataless and Prompt on our own. From our experiments, we observe that the numbers reported in \citet{van2022don} is much lower than our reimplemented prompt-based zero-shot learning results, for reasons unknown to us.

For \emph{transfer-learning based zero-shot inference} methods, we use the same verbalizer as {\ours} and the prompt template provided  from the authors for inference with the released pretrained models. 

For \emph{weakly-supervised learning} and \emph{zero-shot dataset generation}  methods, we use the code released by the authors with the optimal hyperparameters reported in the corresponding paper if available.  
As the code for~\cite{van2022don} is \textbf{not publicly available}, we reimplement this method based on the information from the paper.
If fine-tuning is involved, we use the same pretrained RoBERTa-base as the classifier $C_{\phi}$ with the label smoothing strategy for fair comparison. 

\subsection{Implementation Details for {\ours}}
Table~\ref{tab:hyperpara} lists the hyperparameters used for {\ours}.
Note that we keep them  \emph{same} across all tasks without any further tuning. Under the zero-shot learning setting, there is \emph{no validation set} available. 
For each task, we follow \cite{ye2022zerogen} to use a portion (\eg, 10\%) of the pseudo dataset as the validation set for model selection.
If the total number of the training data for a specific category exceeds 3000, we randomly sample a subset with 3000 samples for that category.

\subsection{Number of Parameters in {\ours}} 
The retrieval model $R_{\theta}$ uses \texttt{BERT-base-uncased} as the backbone with 110M parameters, and the classification model $C_{\phi}$ uses \texttt{RoBERTa-base} as the backbone with 125M parameters.
\subsection{Computation Environment} 
All experiments are conducted on \emph{CPU}: Intel(R) Core(TM) i7-5930K CPU @ 3.50GHz and \emph{GPU}: NVIDIA GeForce RTX A5000 GPUs using python 3.8 and Pytorch 1.10. 

\begin{table}[t]
\centering
\resizebox{1\columnwidth}{!}{
\begin{tabular}{lll}
\toprule
\textbf{Task} & \textbf{Template ID}  & \textbf{Verbalizers}  \\
\midrule
\multirow{4}{*}{\textbf{AGNews}} & \#0 (Original) &  politics, sports, business, technology  \\
& 1 &  world, football, stock, science  \\
& 2 &  international, basketball, financial, research  \\
& 3 &  global, tennis, profit, chemical  \\
\midrule
\multirow{4}{*}{\textbf{Sentiment}} &  \#0 (Original) &  great, bad  \\
& 1 &  good, awful  \\
& 2 &  awesome, terrible  \\
& 3 &  incredible, horrible  \\
\bottomrule
\end{tabular}
}
\caption{
Different verbaliers used for expriments in section \ref{sec:verbalizers}. 
}
\vspace{-1ex}
\label{tab:extra_prompts}
\end{table} 
\begin{table*}[t]
\centering
\resizebox{1.8\columnwidth}{!}{
\begin{tabular}{l*{6}{c}}
\toprule
\bf Method/Dataset & \bf IMDB & \bf SST-2 &\bf Rotten Tomato & \bf Elec &\bf Yelp & \bf Avg. \\ 
\midrule
{Prompting}* & 77.31 & 82.63 & 78.66 & 78.03 & 80.30 & 79.39 \\
 ZeroGen*~\cite{ye2022zerogen} & 80.41& 82.77  & 78.36 & 85.35 & 87.84 & 82.94  \\
ProGen*~\cite{ye2022progen} & 84.12 & 87.20 & 82.86 & 89.00 & 89.39 & 86.51 \\
SuperGen~\cite{meng2022generating} & 84.58	& 86.70 &	79.08 &	90.58 &	89.98 & 86.18 \\
Mining~\cite{van2022don}&77.36 &	80.73 &	76.73 & 	85.87 &	90.36 &	82.21 \\ \midrule
\bf \ours &87.84 &	85.32 &	81.42	 &89.83	 &89.00 & \bf	86.68\\
\bottomrule
\end{tabular}
}
\caption{
Results with recent baselines using DistillBERT~\cite{sanh2019distilbert} as $C_{\phi}$. *: Results are copied from the previous papers~\cite{ye2022progen,ye2022zerogen}. 
}
\vspace{-1ex}
\label{tab:zerogen}
\end{table*}

\section{Additional Information on Experiments Setups}
\subsection{Setup for Fine-tuning $C_{\phi}$ with Few Labeled Examples}
\label{sec:few_apd}
Under the few-shot setting, we follow \cite{meng2022generating} to split the data into two parts: half the data as training set, and the remaining as the validation set. 
When a few labeled samples are available, we first fine-tune the classifier $C_{\phi}$ on the few-shot training set (denoted as $C_{\phi}^{\text{init}}$), and  use $C_{\phi}^{\text{init}}$ to remove the noisy instances with the method in Eq.~\ref{eq:filter} for both our method and baselines. 
Then, we continue fine-tuning the classifier on the generated data.

\subsection{Setup for Zero-shot Learning with Different Verbalizers}
\label{sec:prompt_apd}
We list the set of verbalizers used for section \ref{sec:verbalizers} in table \ref{tab:extra_prompts}. 

\subsection{Setup for Large Language Models for  Verbalizer Expansion}
\label{sec:llm_expan}
For verbalizer expansion, we use GPT-4~\cite{openai2023gpt4} as the LLM backbone, and the prompt format is shown in the followings:

\fbox{
  \begin{minipage}{7cm} 
    \texttt{Suppose you are asked to perform text classification with the following labels. Can you generate 10 relevant keywords for each of the categories?}
  \end{minipage}
}

By inputting the verbalizers of each class into the chatbox, the LLM can output a series of keywords to enrich the verbalizer. After obtaining the keywords, we manually remove keywords that occur in more than one category, and the remaining keywords will be used for retrieval.

\section{Comparison with Recent Baselines}
\label{sec:compare_zerogen}

We provide additional empirical studies to compare {\ours} with some recent works. 
As \cite{ye2022progen,ye2022zerogen} use a smaller PLM, namely DistillBERT~\cite{sanh2019distilbert} for their experiments, we use the same DistillBERT encoder to finetune our model and several baselines (\eg Mining~\cite{van2022don} and SuperGen~\cite{meng2022generating}). The result is shown in table~\ref{tab:zerogen}.

Overall, we observe that {\ours} outperforms most of  these baselines with  DistillBERT as the classifier. It achieves competitive performance with ProGen, which relies on several additional techniques including influence estimation, multi-round in-context feedback from a billion-scale language model, and noise-robust loss functions. Note that these techniques are orthogonal to our method, and can be potentially integrated with {\ours} for better performance.

\begin{table*}[t]
\centering
\renewcommand\arraystretch{0.9}
\resizebox{1.9\columnwidth}{!}{
\begin{tabular}{l*{5}{c}}
\toprule
\bf Task & {\textbf{Datasets}} & \bf Performance of {\ours} & \bf Fully-supervised Performance & \bf $\Delta$ Performance Gap & \bf Lexical Similarity \\ 
\midrule
\multirow{4}{*}{Topic} 
& AG News   & 85.0 & 94.6 & 10.0\% & 0.427 \\
& DBPedia   & 87.6 & 99.2 & 11.2\% & 0.566\\
& Yahoo     & 59.4 & 76.8 & 17.4\% & 0.362 \\
& NYT       & 74.5 & 88.2 & 15.5\% & 0.530\\
\midrule
\multirow{5}{*}{Sentiment} 
& IMDB   & 89.9 & 94.4 & 4.5\% & 0.497 \\
& MR     & 82.5 & 91.3 &  8.8\% & 0.306 \\
& SST-2  & 88.9 & 96.2 & 7.3\% & 0.296 \\
& Amazon & 92.3 & 95.4 & 3.1\% & 0.714 \\
& Yelp   & 93.0 & 97.2 & 4.2\% & 0.408 \\
\bottomrule
\end{tabular}
}
\vspace{-1ex}
\caption{
The detailed value for the performance gap and the lexical similarity between the task-specific corpus and the general-domain corpus $\cD$. 
}
\label{tab:performance_gap}
\end{table*}

\section{More Details on Performance Gaps and Lexical Similarities}
\label{sec:jaccard_apd}
\subsection{Calculating the  Similarity between the Corpus and Target Tasks}
We use the weighted Jaccard similarity $J(T, \cD)$ to measure distrbution similarities  between the corpus $\cD$ and the target task $T$, described as follows: 
Denote $C_k$ as the frequency of word $k$ in the corpus $\cD$ and $T_k$ for the target task $T$ respectively. The weighted Jaccard similarity $J(T, \cD)$ is defined as:
\begin{equation}
\setlength{\abovedisplayskip}{5pt}
\setlength{\belowdisplayskip}{5pt}
J(T, \cD)=\frac{\sum_k \min \left(C_k, T_k\right)}{\sum_k \max \left(C_k, T_k\right)},
\end{equation}
where the sum is over all unique words $k$ present in $\cD$ and $T$.

\subsection{The Performance Gap and Lexical Similarity for All Datasets}
The details for the performance gap as well as the lexical similarity to the general-domain corpus are shown in Table \ref{tab:performance_gap}.

\begin{figure*}[!t]
\centering
\includegraphics[width=0.99\linewidth]{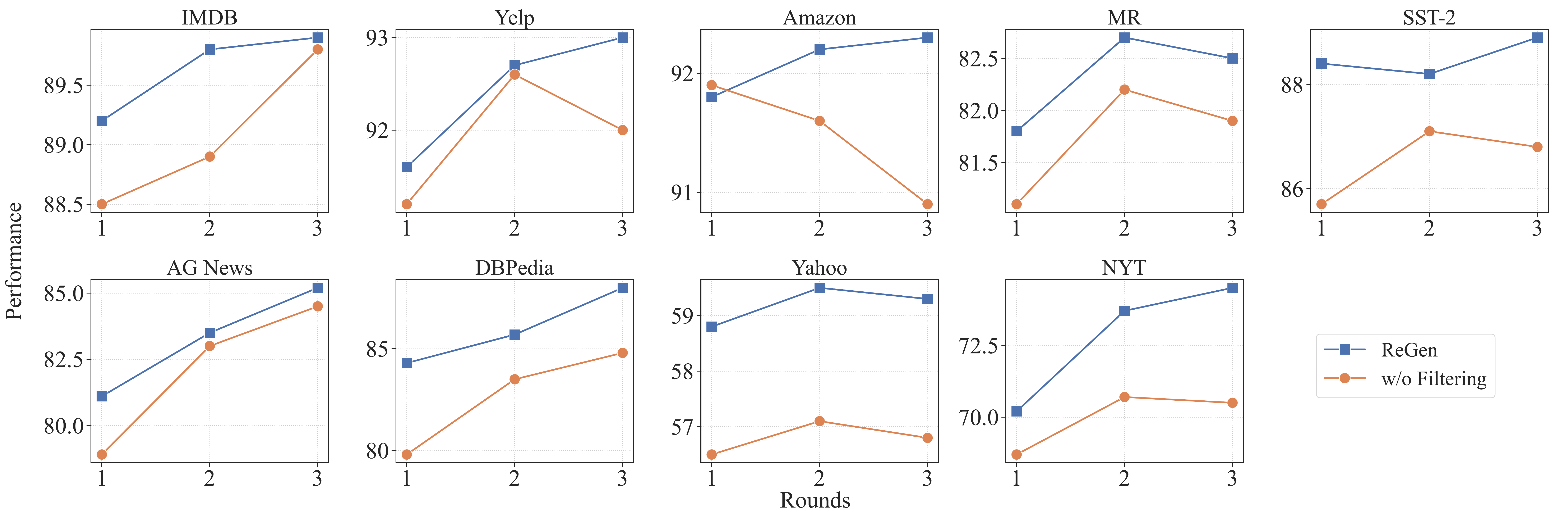}
\vspace{-1ex}
\caption{Effect of filtering, per task results.
}
\label{fig:filtering_per_task}
\end{figure*}

\begin{figure*}[!t]
\centering
\includegraphics[width=0.99\linewidth]{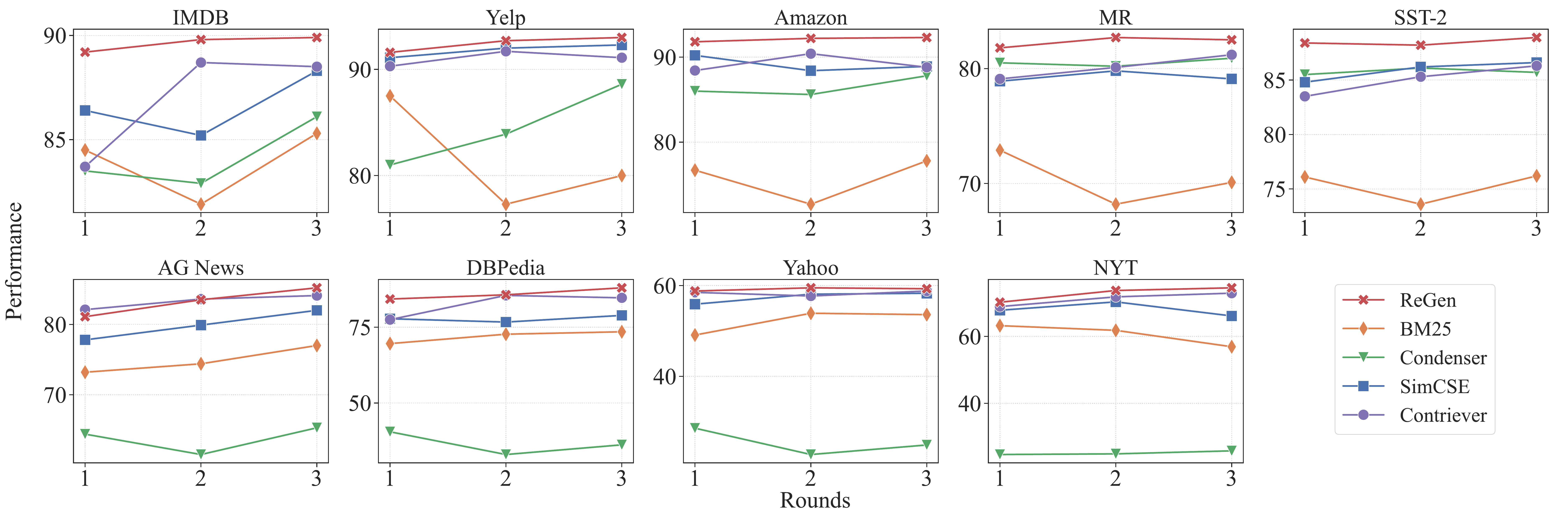}
\vspace{-1ex}
\caption{Comparisons of different dense retrieval models, per task results.
}
\label{fig:dr_model_per_task}
\end{figure*}

\begin{figure*}[!t]
\centering
\includegraphics[width=0.99\linewidth]{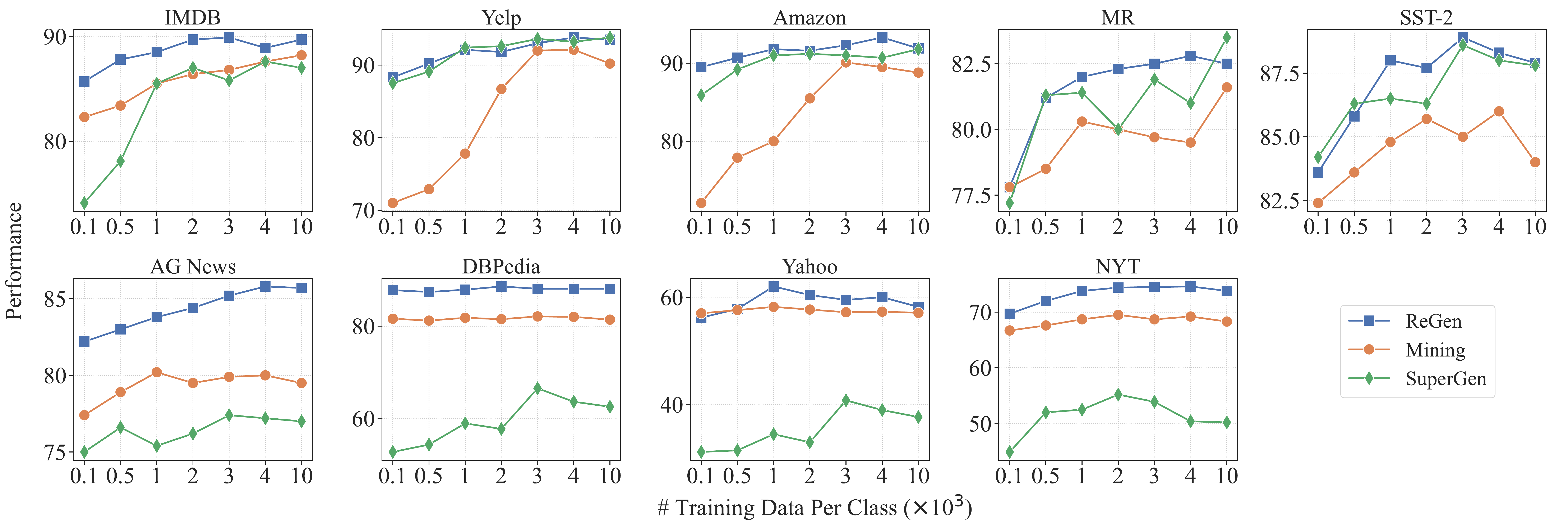}
\vspace{-1ex}
\caption{Performance of the different amount of training data, per task results.
}
\label{fig:training_data_per_task}
\end{figure*}
\begin{figure*}[!t]
\centering
\includegraphics[width=0.8\linewidth]{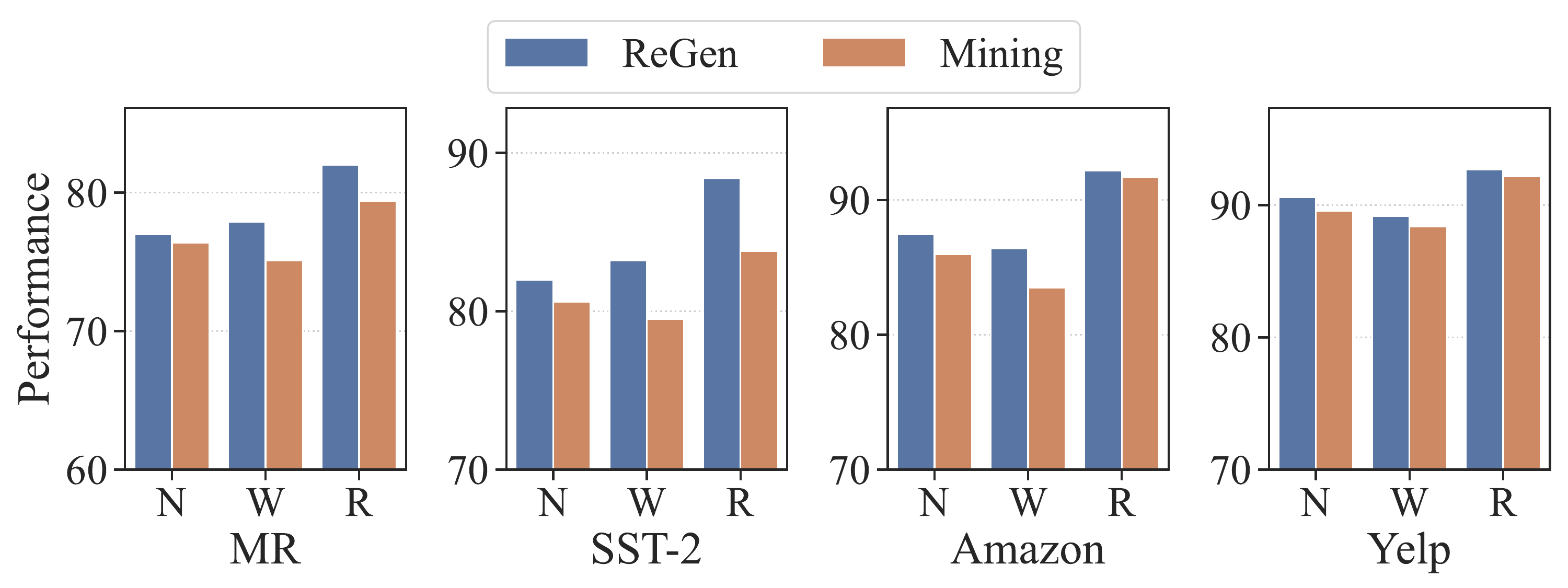}
\vspace{-1ex}
\caption{Performance of {\ours} using different subsets of corpus on other sentiment classification tasks.
}
\vspace{-2ex}
\label{fig:subset_per_task}
\end{figure*}

\section{Additional Per-task Results}
\label{sec:per_task_results}
We show the results for each task in this section.
Specifically, we present the performance of {\ours} and its variation of without the filtering step in Fig.~\ref{fig:filtering_per_task};
we present the performance of {\ours} with different dense retrieval models as $R_{\theta}$ in Fig.~\ref{fig:dr_model_per_task};
we illustrate the performance under different volume of training data for {\ours} and baselines in Fig.~\ref{fig:training_data_per_task};
we demonstrate the effect of different corpus $\cD$  on the final performance in   Fig.~\ref{fig:subset_per_task}.
\begin{table}[t]
\centering
\renewcommand\arraystretch{0.95}
\resizebox{\columnwidth}{!}{
\begin{tabular}{l*{4}{c}}
\toprule
\bf Dataset & {\textbf{Verbalizer Group}} & \bf Mining & \bf SuperGen & \bf{\ours} \\ 
\midrule
\multirow{5}{*}{Yelp} 
&\# 0 (Original) & 92.3	 &\bf 93.6 &	93.0\\
&\# 2  & 85.4 &	91.6 &	\bf 91.9 \\
&\# 3  & 93.4 &	91.2	 &\bf  94.5\\
&\# 4  & \bf 93.2 &	\bf 93.2	 & 92.8 \\
\cmidrule(l){2-5} 
& Avg. $\pm$ Std. & 91.1$\pm${3.8} & 92.4$\pm${1.2} & \bf 93.1$\pm${1.1}\\
\midrule
\multirow{5}{*}{Amazon} 
&\# 0 (Original) & 92.0 &	91.0 &\bf	92.3 \\
&\# 1 & 86.8 &	90.6 &	\bf91.0 \\
&\# 2 & 91.4 &	88.9 &\bf	93.1 \\
&\# 3 & 90.7 &	91.5 &\bf	92.0 \\
\cmidrule(l){2-5} 
& Avg. $\pm$ Std. &  90.2$\pm$2.3 & 90.5$\pm$1.1 & \bf 92.1$\pm$0.8 \\
\midrule
\multirow{5}{*}{MR} 
&\# 0 (Original) & 79.7	& 81.9 &\bf	82.5 \\
&\# 1 & 79.5 &	80.8 &\bf	83.6 \\
&\# 2 & 82.3 &	79.1 &\bf	85.2 \\
&\# 3 & 81.6 &	82.2 &\bf	83.1 \\
\cmidrule(l){2-5} 
& Avg. $\pm$ Std. &  80.8$\pm$1.3 & 81.0$\pm$1.4 & \bf 83.6$\pm$1.2 \\
\midrule
\multirow{5}{*}{SST-2} 
&\# 0 (Original) & 85 &	88.6 & \bf	88.9 \\
&\# 1 & 84.2 &	86.6 &	\bf88.2 \\
&\# 2 & 87.8 &	85.4 &\bf	89.5 \\
&\# 3 & 86.7 &	86.8 &\bf	88.4 \\
\cmidrule(l){2-5} 
& Avg. $\pm$ Std. &  85.9$\pm$1.6 & 86.8$\pm$1.3 & \bf 88.8$\pm$0.6 \\
\bottomrule
\end{tabular}
}
\caption{
Results with different verbalizers on other sentiment analysis datasets.
}
\vspace{-1ex}
\label{tab:prompt_study_extra}
\end{table}

\begin{table}[t]
\centering
\renewcommand\arraystretch{0.96}
\resizebox{\columnwidth}{!}{
\begin{tabular}{l*{4}{c}}
\toprule
\bf Dataset & {\textbf{Metrics}} & \bf Mining & \bf SuperGen & \bf{\ours} \\ 
\midrule
\multirow{3}{*}{Sentiment} & Correctness $(\uparrow)$ & 0.815 &	0.971 & \bf	0.986 \\
& Diversity $(\downarrow)$ & \bf 0.144 &	0.915 &	0.361 \\
& Distribution Sim. $(\uparrow)$ &0.856	 & 0.803 & \bf	0.865\\
\midrule
\multirow{3}{*}{AG News} & Correctness $(\uparrow)$ & 0.746 & 0.649 & \bf 0.805 \\
& Diversity $(\downarrow)$ & \bf 0.117 & 0.818 & 0.330 \\
& Distribution Sim. $(\uparrow)$  & \bf 0.799 & 0.687 & 0.686 \\
\midrule
\multirow{3}{*}{DBPedia} & Correctness $(\uparrow)$ & 0.791 & 0.516 & \bf 0.909 \\
& Diversity $(\downarrow)$ & \bf 0.223 & 0.765 & 0.377 \\
& Distribution Sim. $(\uparrow)$  & 0.874 & 0.662 & \bf 0.920 \\
\midrule
\multirow{3}{*}{NYT} & Correctness $(\uparrow)$ & 0.730 & 0.811 & \bf 0.893  \\
& Diversity $(\downarrow)$ & \bf 0.100 & 0.717 & 0.342  \\
& Distribution Sim. $(\uparrow)$  & 0.511 & \bf 0.643 & 0.622  \\
\midrule
\multirow{3}{*}{Yahoo} & Correctness $(\uparrow)$ & 0.771 & 0.518 & \bf 0.832 \\
& Diversity $(\downarrow)$ & \bf 0.089 & 0.768 & 0.335  \\
& Distribution Sim. $(\uparrow)$  & \bf 0.810 & 0.602 & 0.797 \\
\bottomrule
\end{tabular}
}
\caption{
Automatic evaluation results on all datasets. Note that we only generate one dataset for all sentiment analysis tasks.
}
\label{tab:auto_eval_extra}
\end{table}
 
Besides, in table~\ref{tab:prompt_study_extra} we illustrate the performance of {\ours} and baselines on all sentiment analysis datasets; in table~\ref{tab:auto_eval_extra}, the automatic evaluation results for all datasets are shown.

\section{Details for Quality Analysis}
\subsection{Automatic Evaluation}
\label{sec:autoeval_apd}
We provide the details for automatic measurements of the dataset quality as follows.

For \emph{correctness}, we first fine-tune a RoBERTa-Large model on the original dataset\footnote{For sentiment analysis, we combine the training set of five datasets together as the final training set.},  and use the fine-tuned model as an oracle to evaluate the correctness of the synthetic dataset.  

For \emph{diversity}, we use the self-BLEU~\cite{selfbleu}, which computes the BLEU-4 score of each generated text with other generations in the dataset as references, as the metric.  
Note that for self-BLUE, a \emph{lower} score implies higher diversity.

Besides, we use MAUVE~\cite{pillutla2021mauve} with the default hyperparameter settings to measure the \emph{distribution similarity}. 
MAUVE is originally proposed for comparing the learnt distribution of a text generation model and the distribution of human-written text, and we adapt MAUVE to measure the similarity between the distribution of the synthetic dataset and the real dataset.
A higher value indicates that the distribution of the synthetic dataset and the real dataset is closer, thus the quality of the synthetic dataset is higher. 

\subsection{Human Evaluation}
\label{sec:human_eval_apd}

Apart from the automatic evaluation, we also perform human evaluation to manually evaluate the quality of the synthetic dataset. 
We ask four volunteer students from our institute (apporved by the ethics review board) for participation. 
For human evaluation, the evaluation form is listed as below.
\begin{itemize}
    \item \textbf{Correctness}: Whether the text is relevant to the corresponding label?
    \begin{itemize}
        \item 2: Accurate: The content is accurate for the label.
        \item 1: Related: The content is related but not accurate for the label.
        \item 0: Not relevant: The content is not relevant to the label.	
    \end{itemize}
    \item \textbf{Informativeness}: Whether the text is fluent and similar to human-generated text?
    \begin{itemize}
        \item 2: Very Informative: The text is very informative and similar to human generated text.
        \item 1: Partially Informative: The text is partially informative and somewhat close to human generated text.
        \item 0: Not Informative: The text is not fluent/informative at all.		
    \end{itemize}
    \item \textbf{Diversity}: Whether the generated text within each class is diverse enough?
    \begin{itemize}
        \item 2: Diverse: The documents cover different topics related to the label.
        \item 1: Partially Diverse: There exists some overlapping documents.
        \item 0: Not Diverse: All the documents look similar.
    \end{itemize}
\end{itemize}

\section{Case Studies}
\label{sec:case_study}
We present some examples of the curated dataset in the table \ref{tab:case_studies_sentiment} and \ref{tab:case_studies_agnews}. 
Note that \emph{filtered} means the data is first retrieved by $R_{\theta}$ but is later identified as incorrect sample by the classifier. 
Overall, we observe that the dataset of SuperGen contains similar sentences across samples (\eg \texttt{a great example of the kind of movie} for sentiment analysis datasets), and the mining-based approach often contains documents that are less informative (\eg \texttt{World famous hapuna beach is just minutes away} for topic classification). 
In contrast, {\ours} goes beyond the exact keyword matching and can retrieve diverse and informative documents. Moreover, the noisy samples can be filtered through self-consistency regularization. 
These cases corroborate the result on both automatic and human evaluation, and justify the higher quality of the dataset curated by us.
\begin{table*}[!t]
\centering
\small 
\resizebox{\textwidth}{!}{
\begin{tabular}{lll}
\toprule
\textbf{Method}& \textbf{Label} & \textbf{Generated Text} \\
\midrule
\multirow{6}{*}{\textbf{SuperGen}} & \blue{positive} & {The film is a great example of the kind of movie that you can watch over and over.} \\
& \blue{positive} & {The movie was very good and it had a lot of action in it. I would recommend this to anyone who likes action..} \\
& \blue{positive} &  {This film is a great example of the kind of movie that you can watch with your kids and not have to worry about anything inappropriate.} \\ 
\cmidrule(l){2-3}
& \red{negative} & {The film was a total waste of time. I would not recommend this movie to anyone.} \\
& \red{negative} & {This film is a waste of time. It has no plot and the acting was terrible. I would not recommend this movie to anyone.} \\
& \red{negative} & {This movie is not worth the time or money to watch it. It was a waste of my time.} \\
\midrule
\multirow{6}{*}{\textbf{Mining}} &  \blue{positive} & {The dvd arrived very quick..} \\
&  \blue{positive} & {I can't stop playing them right now.} \\
& \blue{positive} &  {I cannot use it on the hardwood floors because i am afraid water might get down under them (they are very old  but have been refinished.).} \\ 
\cmidrule(l){2-3}
&  \red{negative} & {The plastic handle is not able to be taken apart so i don't know where the leak was exactly coming from.} \\
&  \red{negative} & {Don't know this for sure, but it seems likely.} \\
&  \red{negative} & {OK, this cd makes me sad.} \\
\midrule
\multirow{12}{*}{\textbf{\ours}}& \multirow{2}{*}{\blue{positive}} & {Great I bought this toy for my son's 3rd birthday and only after 2 months he now sings the alphabet song all the time. It is a great education} \\ 
& & {toy and also very durable.} \\
&  \multirow{2}{*}{\blue{positive}} & {After seeing the movie “12 Years A Slave,” I wanted to read the book. The experience of watching the movie drew me into the story of} 
\\
& & {Solomon Northup’s life.} \\
&  \multirow{2}{*}{\blue{positive}} &  {This is a must see film for all ages I would have given this film 10 stars if they would have let me. This is one of those films that somehow} \\ 
& & {got overlooked in the theaters..} \\ 
&  \multirow{2}{*}{\blue{positive} \purple{(filtered)}} &  Excellent but still not Perfect. Don't take my title or rating the wrong way. My experience with the first 2 Harry Potter Movies have been  \\
& &  excellent, but in the 2nd movie, the Chamber of Secrets, A lot of parts were taken out...  \\ 
\cmidrule(l){2-3}
&  \red{negative} & {Worst movie ever A good example of what is wrong with Hollywood today. I have never looked at my watch more times during a movie.} \\
&  \red{negative} & {Bad book I did not like it. It is a bad story. Wolfs are not bad like in the story. Peter doesnt listen to his grandpa, so it is a bad example.} \\
&  \red{negative} & {Silicon Valley... I do not like this game. The directions are hard to follow and I did not like the graphics at all.} \\
&  \red{negative} \purple{(filtered)} & {how can people dislike this charming film, this is very wonderful film that works for both audlts and kids.} \\
\bottomrule
\end{tabular}
}
\caption{
Example retrieved texts of {\ours} and two baselines on synthetic dataset for sentiment analysis.
}
\label{tab:case_studies_sentiment}
\end{table*}

\begin{table*}[!t]
\centering
\small 
\resizebox{\textwidth}{!}{
\begin{tabular}{lll}
\toprule
\textbf{Method}  & \textbf{Label} & \textbf{Generated Text} \\
\midrule
\multirow{9}{*}{\textbf{SuperGen}}  
& \blue{politics} & {The opinions expressed in this commentary are solely those of John Avlon..} \\
& \multirow{2}{*}{\blue{politics}} & {TL;DR Correction of Sept 30 article on Pres Bush's visit to New York City, which misstated his role in campaign finance reform legislation that was signed into} \\
& &  law by Gov George Pataki. \\
\cmidrule(l){2-3}
&  \orange{sports} & {TL;DR Correction of Nov 12 article on New York Yankees pitcher Roger Clemens regarding his use of steroids; he is not using steroids and has never used them.} \\
&  \orange{sports} & {TL;DR Correction of Aug 25 article on New York Yankees player Mariano Rivera regarding his role in team's World Series victory over Arizona Diamondbacks.} \\
\cmidrule(l){2-3}
&  \green{business} & {The company said it had hired the law firm of Paul, Weiss, Rifkind, Wharton \& Garrison to conduct an independent investigation.} \\
&  \green{business} & {The company said it had hired the law firm of Debevoise \& Plimpton to conduct an independent investigation.} \\
\cmidrule(l){2-3}
&  \red{technology} & {TL;DR The National Science Foundation awarded \$32 million to the University of California, Berkeley, for research on how people use computers in their  lives.} \\
&  \red{technology} & {TL;DR The New York Times Magazine publishes its annual list of the 100 most influential people in science, technology, engineering or math.} \\
\midrule
\multirow{9}{*}{\textbf{Mining}}  
& \blue{politics} & {World famous hapuna beach is just minutes away.} \\
& \blue{politics} & {At the same time, we should not let our good fortune make us callous to the effect of suffering on most of the world population.} \\
\cmidrule(l){2-3}
&  \orange{sports} & {According to multiple sportsbooks, curry isn't even in the top-five likeliest mvp candidates for 2016-17.} \\
&  \orange{sports} & {Sky sports reported tonight chelsea have held talks over the former napoli manager's future.} \\
\cmidrule(l){2-3}
&  \green{business} & {I am not starry-eyed about the news business 2014 and it is a business.} \\
&  \green{business} & {Fostering a sense of autonomy amongst employees should be a central goal for all business leaders.} \\
\cmidrule(l){2-3}
&  \multirow{2}{*}{\red{technology}} & {Notebook casing supplier catcher technology was forced to close one facility over environmental concerns, while iphone supplier pegatron was fined for spewing} \\
& & {harmful gases during the manufacture of products.} \\
&  \red{technology} & {Panaji: goa police in association with a bengaluru-based start-up has come up with a technology which can detect unauthorized drones.} \\
\midrule
\multirow{24}{*}{\textbf{\ours}}  
& \multirow{2}{*}{\blue{politics}} & {The United Nations Human Rights Commissioner Navi Pillay has called for an international probe into war crimes committed in Sri Lanka during the final stages} \\
& &  {of its ethnic conflict, according to a media report on Sunday.} \\
& \multirow{2}{*}{\blue{politics}} & {Police in Bolivia have rebelled against the government, abandoning their posts and marching through the streets along with protesters. It's a sign of growing anger} \\
& & {over alleged voter fraud in last month's election. Protests since the poll have resulted in three deaths.} \\
& \multirow{2}{*}{\blue{politics} \purple{(filtered)}} & {An Australian in ASEAN. It sounds like the title of an innocent-abroad movie: the hero has adventures, blunders and embarrasses. But in the end Aussie charm} \\
& & {and grit prevail; romance blossoms and the outsider becomes an insider..} \\
\cmidrule(l){2-3}
& \multirow{2}{*}{\orange{sports}}  & {Tom Brady and Bill Belichick likely will go down as the greatest quarterback/coach combo in NFL history, especially after winning their fifth Super Bowl} \\
& & {together with a thrilling 34-28 overtime victory against the Atlanta Falcons in Super Bowl LI on Sunday night.} \\ 
&  \multirow{2}{*}{\orange{sports}}  & {Manchester City's quest for four trophies continued with a 5-0 thrashing of Burnley to march into the FA Cup fifth round as League One Shrewsbury narrowly} \\
& &  missed out on shocking Wolves in a 2-2 draw on Saturday. \\
&  \multirow{2}{*}{\orange{sports} \purple{(filtered)}}  & {The growing scandal involving the new designer steroid THG gives sports fans one more thing other than sports to worry over. To be a sports fan is to get a} \\ 
& & {constant education in subjects that don't necessarily interest you.} \\
\cmidrule(l){2-3}
&  \multirow{2}{*}{\green{business}} & {THE HAGUE, Netherlands, March 14, 2019 /PRNewswire/ -- Royal Dutch Shell plc RDS.A, +0.35\% RDS.B, +0.19\% filed its Annual Report on Form 20-F for} \\
& & {the year ended December 31, 2018, with the U.S. Securities and Exchange Commission.} \\ 
&  \multirow{2}{*}{\green{business}} & {Dimensions International Inc. has acquired Sentel Corp., creating a company that will have more than \$100 million in annual revenue. Terms of the deal were not} \\
& & {disclosed.} \\ 
&  \multirow{2}{*}{\green{business} \purple{(filtered)}} & {Mercosur full members (Argentina, Brazil, Paraguay and Uruguay) rank poorly in the Forbes magazine annual Best Countries for Business, with the best listed,} \\
& & {Chile and Peru, in positions 24 and 42, out of 134 countries surveyed worldwide.} \\ 
\cmidrule(l){2-3}
&  \multirow{2}{*}{\red{technology}} & {SpaceX's next-generation rocket, the Starship, is 50 meters long and powered by three Raptor engines, creating a whopping 12,000 kN of thrust. It is designed to} \\
& & {haul large amounts of cargo and eventually passengers into space, for missions to the moon and potentially to Mars and beyond as well. } \\ 
&  \multirow{2}{*}{\red{technology}} & {Physicians that use the clinical reference tool, DynaMedTM from EBSCO Health, can now access the valuable, evidence-based content anywhere with the new} \\
& & {DynaMed mobile app. The new app has been redesigned to make it easier and faster for physicians to find answers to clinical questions.} \\ 
&  \multirow{2}{*}{\red{technology} \purple{(filtered)}} & {Cookson is science editor at the FT. He joined the newspaper in 1988 as technology editor and has also written about the chemical and pharmaceutical industries.} \\
& & {Previously, he was the science and medical correspondent for BBC Radio.} \\ 
\bottomrule
\end{tabular}
}
\caption{
Example retrieved texts of {\ours} and two baselines on the synthetic dataset for AG News.
}
\label{tab:case_studies_agnews}
\end{table*}

\begin{table*}[t]
\centering
\small 
\resizebox{\textwidth}{!}{
\begin{tabular}{lll}
\toprule
\textbf{Round}& \textbf{Label} & \textbf{Generated Text} \\
\midrule
\multirow{4}{*}{\textbf{1}} & 
\multirow{2}{*}{\blue{positive}} & {"Deceptions" was one of the best films I have seen in a long time. Stefanie Powers was excellent as Sabrina and Samantha. } \\
& & {The rest of the cast was also very good.} \\ \\
& \multirow{1}{*}{\red{negative}} & {I honestly have no idea what to say about this movie. It literally left me speechless….in a very, very not-good way.} \\
\midrule
\multirow{5}{*}{\textbf{2}} & 
\multirow{2}{*}{\blue{positive}} & {I saw the film last weekend and enjoyed it. From the point of view of movie craftsmanship, it's hard to go wrong with the} \\
& & {talent combination of Steven Spielberg, Meryl Streep, Tom Hanks, and John Williams.} \\ \\
& \multirow{2}{*}{\red{negative}} & {To be frank, it is a really bad movie. The cheap symbolism would make a junior high English teacher blush (including the} \\
& & {title), and the lopsided view of racism in America was painfully and repeatedly portrayed.} \\
\midrule
\multirow{6}{*}{\textbf{3}} & 
\multirow{2}{*}{\blue{positive}} & {"Letting Go," with Sharon Gless and John Ritter, was a warm, funny and dramatic movie. I loved it. It was a fresh and } \\
& & {wonderful romance.} \\ \\
& \multirow{3}{*}{\red{negative}} & {First of all, I would like to say that I think the movie did an excellent job of following the events in the book. But they did} \\
& & {a pretty bad job of leaving some crucial parts out of the movie. In the book, you get a pretty strong sense of the bond and} \\
& & {relationship between the characters. In the movie, you don’t really see that bond at all.} \\
\bottomrule
\end{tabular}
}
\caption{
Example retrieved texts of {\ours} over three rounds for sentiment datasets.
}
\label{tab:case_studies_multi_review}
\end{table*}

\begin{table*}[t]
\centering
\small 
\resizebox{\textwidth}{!}{
\begin{tabular}{lll}
\toprule
\textbf{Round}& \textbf{Label} & \textbf{Generated Text} \\
\midrule
\multirow{11}{*}{\textbf{1}} & 
\multirow{2}{*}{\blue{politics}} & {The UN voiced hope Monday that a meeting this week of a committee tasked with amending Syria's constitution can} \\
& & {open the door to a broader political process for the war-ravaged country.} \\ \\
& \multirow{2}{*}{\orange{sports}} & {LaLiga may boast football superpowers Real Madrid and Barcelona but the league is keen to help other Spanish sports} \\
& & {succeed too.} \\ \\
& \multirow{2}{*}{\green{business}} & {Corporate America is slowly starting to give cash back to investors with dividends and buybacks. Companies are also} \\
& & {spending cash on mergers.} \\ \\
& \multirow{2}{*}{\red{technology}} & {Google said on Wednesday it had achieved a breakthrough in research, by solving a complex problem in minutes with} \\
& & {a so-called quantum computer that would take today's most powerful supercomputer thousands of years to crack.} \\
\midrule
\multirow{12}{*}{\textbf{2}} & 
\multirow{2}{*}{\blue{politics}} & {The death toll in Eastern Ghouta stands at nearly 500, and it remains unclear how the sustained bombing campaign in} \\
& & {the region will stop—despite a UN vote.} \\
\\
& \multirow{2}{*}{\orange{sports}} & {Barcelona continued their quest to win La Liga with a comfortable 3-0 victory over Leganes yesterday. Luis Suarez} \\
& & {ended his goal drought with a brilliant brace before summer signing Paulinho got on the scoresheet late on.} \\ 
\\
& \multirow{3}{*}{\green{business}} & {For many American companies today it is almost as is the recession never happened as executive incomes rise above} \\
& & {pre-recession levels. According to Standard \& Poor's 500 the average income of an executive in 2010 was \$9 million.}  \\
& & That is 24 percent higher than it was the year prior. \\
\\
& \multirow{2}{*}{\red{technology}} & {Scientists claimed Wednesday to have achieved a near-mythical state of computing in which a new generation} \\
& & {of machine vastly outperforms the world's fastest super-computer, known as "quantum supremacy"} \\
\midrule
\multirow{13}{*}{\textbf{3}} & 
\multirow{3}{*}{\blue{politics}} & {The UN's ceasefire in Syria's rebel-held enclave of Eastern Ghouta was cast into doubt less than 24 hours after the} \\
& & {Security Council voted to uphold it, as residents woke to regime airstrikes and Iran vowed to carry on fighting in} \\
& & {areas it deems held by terrorists.} \\
\\
& \multirow{2}{*}{\orange{sports}} & {Eden Hazard exploded into life and Karim Benzema continued his brilliant scoring run as Real Madrid delivered} \\
& & {another goalfest on Saturday in a 4-0 demolition of Eibar.} \\ 
\\
& \multirow{3}{*}{\green{business}} & {Wall Street's eternally optimistic forecasters are expecting corporate profit growth to surge by the middle of next year} \\
& & {views that are about to collide with reality as hundreds of companies report financial results and update investors on}  \\
& &  their prospects. \\
\\
& \multirow{3}{*}{\red{technology}} & {From ending the opioid epidemic to making fusion power possible, 'Summit' may help researchers meet all sorts of} \\
& & {goals. A \$200-million, water-cooled monster that covers an area the size of two tennis courts, the computer, dubbed} \\
& & {"Summit," has been clocked at handling 200 quadrillion calculations a second. } \\
\bottomrule
\end{tabular}
}
\caption{
Example retrieved texts of {\ours} over three rounds for AG News dataset.
}
\label{tab:case_studies_multi_agnews}
\end{table*}

We also demonstrate the retrieved examples over different rounds in table~\ref{tab:case_studies_multi_review} and \ref{tab:case_studies_multi_agnews}. Note that examples shown in the 2nd and 3rd round are retrieved directly using the concatenation of class-specific verbalizers and document from the previous round. 
The results indicate that {\ours} can iteratively retrieve text that are sementically close to the documents from previous rounds.

\end{document}